\documentclass[11pt, a4paper,  oneside]{report}

\usepackage{float} 

\usepackage{url} 

\usepackage{graphicx}
\graphicspath{ {./images/} }
\usepackage[utf8]{inputenc}
\usepackage[export]{adjustbox}

\usepackage{amsmath}
\usepackage{hyperref}
\hypersetup{colorlinks=True,linkcolor=black,urlcolor=blue,citecolor=black}

\newfloat{fig}{thp}{lof}[chapter]
\floatname{fig}{Figure}
\usepackage{vuwproject}

\title{Feature-based Image Matching for Identifying Individual Kākā}
\author{Fintan O'Sullivan, Kirita-Rose Escott, \\Rachael C.\ Shaw, and Andrew Lensen}

\begin{document}

\begin{abstract}
\noindent This report investigates an unsupervised, feature-based image matching pipeline for the novel application of identifying individual kākā. Applied with a similarity network for clustering, this addresses a weakness of current supervised approaches to identifying individual birds which struggle to handle the introduction of new individuals to the population. Our approach uses object localisation to locate kākā within images and then extracts local features that are invariant to rotation and scale. These features are matched between images with nearest neighbour matching techniques and mismatch removal to produce a similarity score for image match comparison. The results show that matches obtained via the image matching pipeline achieve high accuracy of true matches. We conclude that feature-based image matching could be used with a similarity network to provide a viable alternative to existing supervised approaches. 


\end{abstract}


\maketitle

\tableofcontents


\mainmatter


\chapter{Introduction}\label{C:intro}

Computer vision is a field of artificial intelligence that focuses on using computers to process, analyse and interpret visual data \cite{Nixon2012}. One of the primary tasks of computer vision is image classification, which is the process of categorising images by features extracted from image data \cite{Jensen2017}. Image classification is used for a wide range of tasks including facial recognition, medical imagery, and land use analysis. 

\section{Motivation}

Ecology and conservation is an emerging area for the application of computer vision and image recognition techniques \cite{Weinstein2018} \cite{Christin2019}. Much of the existing work in the field is concerned with image classification of different species \cite{Christin2019}.

Part of the difficulty in applying computer vision techniques in ecology is that the differences between species are so fine-grained. Consequently, identifying individuals of the same species presents an even more complex task for image recognition models than identifying different species. Yet, such models would give ecologists a valuable tool for counting species populations, analysing group behaviour, and observing intraspecies social networking \cite{Weinstein2018}. Currently, ecologists would rely on manual identification practices such as tagging, banding or marking individuals to conduct those studies. Alternatively, researchers could painstakingly manual identify individuals from photos or video footage. Automating that process by developing image classification models that could identify individuals would therefore save time, costs and labour \cite{Silvy2005}. 

The local population of North Island kākā(\textit{nestor meridionalis septentrionalis}) at the Zealandia ecosanctuary in Wellington present an ideal subject for study in the identification of same-species individuals. The unique markings and geometry of their beaks provide high quality discriminating features between individuals that could be used to train image classification models. Initially, there were only fourteen kākā in the kākā population at Zealandia \cite{Zealandia2016}. These kākā were captive-bred, and transferred from other zoos to Zealandia between 2002 and 2007. Since then the kākā population at Zealandia has flourished, and Zealandia has become a source for translocating kākā to other ecosanctuaries in New Zealand, such as the Cape Sanctuary in the Hawke's Bay \cite{Zealandia2016}. 

To monitor the kākā population, kākā born at Zealandia nesting boxes are banded with an identifying colour combination that indicates the individual and the cohort (year) that each kākā was born in. Nevertheless, birds that have been banded can still be difficult to identify. The coloured bands can chip and lose their colour, and many of the colours are difficult to distinguish, particularly under different light conditions. For instance, some kākā have a lime green coloured cohort band that is nearly impossible to distinguish from silver or chipped bands under forest light conditions. Additionally, banding is not a feasible solution to monitoring individuals at large scales. Particularly with coloured banding, there are a finite number of easily identifiable combinations that place a limit on the number of kākā that can be monitored at once.

Since the end of the 2015/16 breeding season, intensive banding operations have been scaled back to only those necessary for research. 750 kākā had been banded by the end of the 2015/16 breeding season and by the end of the 2018/19 breeding season, over 1000 kākā had been banded. 

As the kākā population at Zealandia has grown, the number of unbanded kākā has also grown. Unbanded kākā are hatched outside of the nesting boxes provided at Zealandia in natural cavity nests. These kākā are an equally important subset of the population for researchers to monitor because unbanded kākā will soon make up the majority of the population that frequent Zealandia.  

A machine learning model would offer the ability to continue to monitor individual kākā as the population outgrows the limitations of banding, and help to continue a variety of research projects such as estimating local kākā population size and analysing behaviour patterns in groups of kākā.

\section{Problem Statement}

As discussed, identifying individuals of the same species is a more difficult computer vision task than identifying individuals of different species because there are fewer differences between individuals of the same species. One of the primary implications of this is that an image classification model requires a large dataset for training, containing thousands to millions of images to correctly differentiate between individuals \cite{Miele2021} \cite{Ferreira2020}. If the dataset is too small, it can cause the classification model to rely on temporary or misinformed features. For instance, a model might match images based on individuals positioning rather than content, even though position within an image frame is not really a feature. As another example, in the case of kākā, a minor wound or temporary colouring (due to season or age) might be detected as a key feature for a particular individual, but the model won’t classify the individual the same way a few days or months later. 

Existing attempts to classify individuals have focused on supervised approaches and convolutional neural networks (CNN) \cite{Miele2021} \cite{Ferreira2020}. Supervised approaches learn a function that maps known (labelled) outputs to the input data. The main weakness of these approaches for classifying individuals of the same species is that the models struggle to recognise unknown or unseen individuals. This is because these systems can only predict labels that already exist in the label set (i.e. individuals known to the classifier after training). In the practical application of individual classification systems in ecology, it is expected that new individuals will enter the population over time.  Being able to recognise new individuals and label them accordingly is therefore an important capability of classification systems. Another flaw of the supervised approach in the context of individual classification is that they rely on large datasets of labelled data \cite{Ferreira2020}. Manually labelling datasets of the scale required can be tedious, labour-intensive, and prone to human error \cite{Miele2021}. Indeed, the primary purpose of individual classification models is to replace the need for humans to manually identify individuals from photos. This is all to say that if collecting a large dataset of labelled data was straightforward then an individual identification model would not be necessary. 

\section{Proposed Solution}

We propose an intraspecies individual identification method using an online unsupervised learning approach. Unsupervised learning categorises machine learning tasks in which the outputs (target labels) are unknown and the goal is to discover meaningful patterns in the input data \cite{Celebi2016}. Unsupervised learning approaches are useful when labelled data is not available or difficult or labour intensive to collect \cite{Celebi2016}. Online learning, on the other hand, refers to machine learning methods where the model learns from a sequence of data one at a time \cite{Hoi2021}. Unlike offline training settings, which train a model with the entire training set at once and must be retrained to incorporate new data, online learning does not need to be retrained as new data arrives \cite{Hoi2021}. This means that a model using an online learning approach can handle a growing number of individuals in the population, which we have previously identified as a vital attribute for the practical use of any individual classification system \cite{Hoi2021}. 

In this way, a model using an online unsupervised approach can be thought of as closely mimicking how a human would approach the process of sorting images of animals by individual. As the model receives new images it determines if it has seen the individual in the image before; if it has, the image is placed in the pile of the corresponding individual, and if it has not it begins a new pile. 

\section{Goals}

The goal of this research is to develop an image matching pipeline that could be used to construct a similarity network for images of North Island kākā living at Zealandia. This would follow an online unsupervised approach and allow the use of clustering to identify individuals kākā. This is to be accomplished in five main stages:

\begin{enumerate}
    \item \textbf{Object Localisation:} detect and localise kākā within the kākā image data to filter features for feature extraction.
    \item \textbf{Feature Extraction:} extract interesting and relevant features from the kākā image data.
    \item \textbf{Feature Matching:} match features from multiple images, based on their descriptors.
    \item \textbf{Similarity Scoring:} from the feature matching process, explore different methods for measuring image similarity to formulate a similarity score for images.
    \item \textbf{Image Matching:} find the best match for any given image using the formulated similarity score.
\end{enumerate}

\chapter{Background and Related Work}\label{C:back}

\section{Image Matching}

Computer vision can be thought of as the pursuit of allowing computers to perceive visual information as humans do \cite{Szeliski2022}. To that end, a key part of computer vision is the ability to recognise patterns and understand relationships between multiple images. As a fundamental problem in a range of computer vision tasks such as remote sensing, security and medical imaging, the goal of image matching is to identify patterns in image data and then correspond the similar structure from other images \cite{Szeliski2022}. Since computer vision first became a field, there have been a growing number of methods proposed for image matching, and this number has grown even more with the development of deep learning techniques. 

Regardless of method, image matching comprises primarily of three stages: 

\begin{enumerate}
    \item \textbf{Feature detection:} how the underlying patterns and structures are found
    \item \textbf{Feature description:} how they're described for comparison
    \item \textbf{Feature matching:} how they are matched
\end{enumerate}

Image matching approaches have historically been broadly split into area-based and feature-based approaches, however, deep learning-based techniques are also being developed \cite{Zhang:2018:LTD} \cite{Yi2016} \cite{Laguna2019}.

\subsection{Area-based Image Matching}

The most simple and naive method for image matching is area-based matching. This method directly compares image pixel intensities in windows of a preset size to calculate the similarity between images, rather than detecting any specific image structures \cite{Ma2021}. Area-based matching is sensitive to the window size, the similarity measurement used, as well as image noise and rotation, scale and affine variations \cite{Ma2021}.

\subsection{Feature-based Image Matching}

Feature-based image matching is the more widely used classic approach to image matching \cite{Ma2021} . Feature-based image matching has existed for decades and is more flexible and robust than area-based methods, and therefore has a larger range of applications \cite{Zitova2003}. Feature-based approaches detect the underlying structure of the objects in a given image and use that to construct descriptors that can be used to efficiently compare images \cite{Zitova2003}. Within the feature-based approach, feature detection and feature description are condensed into the one feature extraction stage, which detects suitable keypoints and computes those keypoints' descriptors \cite{Kumar2014}. The effectiveness of the feature-based approach is therefore largely dependent on the feature extraction method. 

An additional method to improve feature-based image matching is to use object detection as part of the feature detection process \cite{Ma2021}. By applying object detection, features are localised to the object which ensures that features are relevant to the object and informative for image matching \cite{Ma2021}. 

\section{Image Segmentation}

Image segmentation is the process of partitioning an image into image segments or regions. It can alternatively be used for object detection, where there is a clear distinction between the object and the background \cite{Toshev2010}. This makes image segmentation a useful tool for object detection in assisting feature-based image matching. Specifically, image segmentation can segment the interest area from the background to assist with feature extraction \cite{Varshney2009}. 

There are three main types of segmentation. Semantic segmentation classifies each pixel in an image as belonging to a specific class \cite{Minaee2022}. Instance segmentation classifies each pixel of a specific class, into the individual instances of that class \cite{Minaee2022}. Panoptic segmentation is a combination of instance and semantic segmentation, classifying each pixel in an image into the class and instance that it belongs to \cite{Minaee2022}. 

Some common image segmentation techniques include k-means clustering, region-based segmentation, thresholding segmentation, and edge-based segmentation \cite{Varshney2009}.

\subsection{Image Segmentation with K-Means Clustering}

K-means clustering is one of the most popular and simple image segmentation methods \cite{Varshney2009}. K-means clustering segments an image into \(k\) regions based on the pixel values in an image \cite{Dhanachandra2015}. The algorithm for k-means clustering is as follows \cite{Likas2003}:

\begin{enumerate}
\item Choose the number of clusters, \(K\)
\item Randomly initialise \(K\) cluster centroids \(C_1, C_2..., C_k\)
\item For each data point \(x\) find the nearest centroid \(C\) and assign \(x\) to that cluster's set of points \(S\)
\item For each set of points \(S_1, S_2,...S_k\), calculate the new centroid \(C_1, C_2..., C_k\) by finding the mean of all points in the set \(S\)
\item Repeat steps 3 and 4 until convergence or until some number of iterations is complete. 
\end{enumerate}

There are three main drawbacks of k-means clustering. Firstly, k-means clustering is very sensitive to the value of \(k\) and the \(k\) value is set before clustering, so good results are dependent on having some prior knowledge to what the best value of \(k\) is for the data \cite{Pham2016}. Secondly, k-means is non-deterministic meaning that results may vary between different runs of the algorithm because the cluster centroids are randomly intialised \cite{Nidheesh2017}. Thirdly, k-means is biased towards spherical segments because it uses centroids to cluster the data \cite{He2022}. This means clusters naturally form circles, with a radius from the centroid to the farthest point in the cluster. Therefore, K-means clustering does not perform well in complex clustering tasks when clusters are in non-spherical shapes \cite{He2022}. 

\section{Feature Extraction}

A crucial component of feature-based image matching is feature extraction, because it involves both feature detection and feature description. Feature extraction is a form of dimensionality reduction and refers to the process of transforming input image data into feature vectors which can be then be used to match images \cite{Nixon2012}. 

Features capture information relevant to the shapes contained within the image and extract important underlying patterns and structures. Features are compared across different images by creating rotation, scale and affine invariant descriptors that can easily be matched against other descriptors.

In the abstract, features can be split into two categories: low-level features and high-level features \cite{Takarli2016}. Low-level features are basic features that can be extracted without information about spatial relationships, such as edges, corners and blobs \cite{Nixon2012}. High-level features are those that can be directly seen and recognised in the image, such as shapes and objects \cite{Nixon2012}. High-level features must have knowledge of spatial relationships within the image \cite{Nixon2012}. Low-level and high-level features can therefore be thought of as local and global features, respectively \cite{Takarli2016}.

Some low-level feature extraction methods include edge detection methods such as Sobel, Canny and Prewitt edge detection, Harris corner detection, and localised feature extraction methods such as SIFT, SURF \cite{Bay2008}, oriented FAST and rotated BRIEF (ORB), binary patterns histogram (LBPH) \cite{Ojala1996}, and histogram of oriented gradients (HoG) \cite{Dalal2005} \cite{Nixon2012}. Localised feature extraction methods can also be considered high-level feature extraction methods because sets of features can be used to characterise and describe objects \cite{Nixon2012}. Other high-level feature extraction methods include template matching, wavelets and Haar wavelets, and generalised and invariant Hough transforms \cite{Nixon2012}. 

\subsection{Scale-invariant feature transform (SIFT)}\label{sec:sift}

SIFT is perhaps the most well-known and widely used feature extraction method for image recognition and object detection. First developed in 1999, SIFT produces features that are scale and rotation invariant which makes them robust to noise and occlusion \cite{Lowe1999}. The SIFT algorithm works in four stages: scale-space extrema detection, keypoint localisation, orientation assignment, and keypoint descriptor creation \cite{Lowe2004}. 

The first stage, scale-space extrema detection, consists of identifying points of interest that are invariant to scale. This is accomplished by searching for features that are stable across different scales, using a continuous function called the scale space. The canonical way to generate a linear scale space is with the Gaussian function, which is a smoothing function that ensures that no new structures are introduced into the image as the scale increases \cite{Lindeberg1994}. The scale space of an image is therefore a function, \(L(x, y, \sigma)\), produced by the convolution of the variable-scale Gaussian function, \(G(x, y, \sigma)\), with an input image, \(I(x,y)\).

 \[ L(x, y, \sigma) = G(x, y, \sigma) * I(x, y)\]
 
 SIFT then finds the difference of Gaussians function, \(D(x,y,\sigma)\), computed from the difference between two scales separated by a constant multiplying factor \(k\), to find keypoints that are stable across different scales. 
 
\begin{align*}
    D(x,y, \sigma) & = (G(x,y,k\sigma) - G(x,y,\sigma)) * I(x,y,) \\
    & = L(x,y,k\sigma) - L(x,y,\sigma) \\
\end{align*}

\begin{figure}[H]
\centering
  \includegraphics[width=0.7\textwidth]{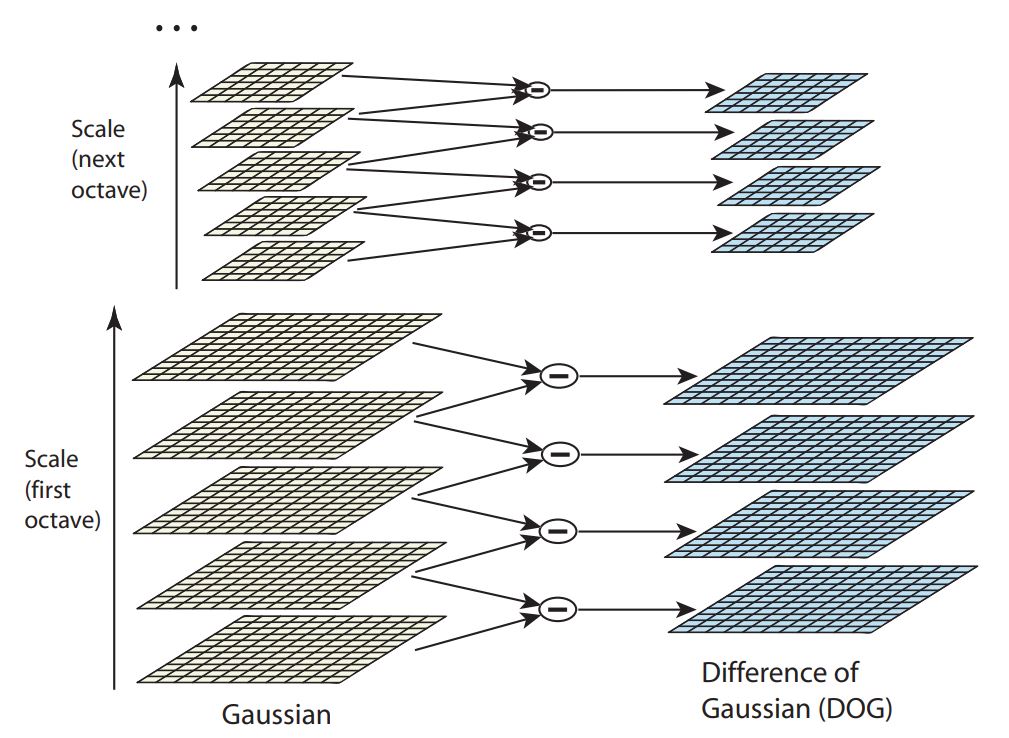}
    \caption{From \cite{Lowe2004}, illustrating how the difference of Gaussians is found.}
  \label{fig:diff-of-gaussians}
\end{figure}

\begin{figure}[H]
\centering
  \includegraphics[width=0.4\textwidth]{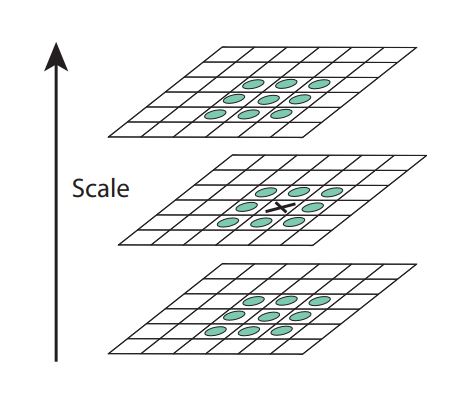}
    \caption{From \cite{Lowe2004}, showing how local extrema are calculated by comparing points with neighbouring points across adjacent scale spaces.}
  \label{fig:local-extrema-calc}
\end{figure}

This is important because the difference of Gaussians provides a good approximation of the Laplacian of Gaussians (LoG), which when normalised with the factor \(\sigma^2\) gives true scale invariance. However, the  \(\sigma^2\) scale normalisation factor is naturally incorporated when the difference of Gaussians differ by a constant scale factor.  So, to generate \(D(x,y,\sigma)\) we can iteratively convolve an initial image, \(I(x,y)\), with Guassians to produce images separated by a constant scale factor \(k\). Each octave (doubling of \(\sigma\)) is further split into \(s\) intervals. Next, each adjacent image scale is used to find the difference of Gaussians, as shown in Figure \ref{fig:diff-of-gaussians}. After each octave, the image scale is down-sampled by a factor of two, and this process is repeated. Finally, to find the local extrema, each point is compared with its eight neighbouring points in the current image and the nine points above and below it in the scale space. If that point is a maxima or minima then it is identified as a keypoint. That process is shown in Figure \ref{fig:local-extrema-calc}.

In keypoint localisation, those keypoints identified in the previous stage are contextualised, and poor keypoints are removed. This includes keypoints that are in low contrast regions of the image or are located close but not on an edge within the images. These keypoints are unstable because they're susceptible to slight noise, so they are discarded.

In orientation assignment, the orientation of each keypoint is computed from local image properties. Keypoint descriptor can then be represented relative to the keypoint's orientation and consequently be invariant with respect to image rotation.  To accomplish this, SIFT takes a region around each keypoint, dependent on the scale, \(L\), at which the keypoint was detected, and measures the gradient magnitude, \(m(x,y)\), and direction, \(\theta(x, y)\) from a set of sample points in that region. 

\begin{align*}
m(x, y) = \sqrt{(L(x + 1, y) - L(x - 1, y))^2 + (L(x, y + 1) - L(x, y - 1))^2} \\
   \theta(x, y) = tan^{-1} (L(x, y + 1) - L(x, y - 1))/(L(x + 1, y) - L(x - 1, y)))
\end{align*}

An orientation histogram with 36 bins, each containing a 10-degree slice of the 360 degrees surrounding the keypoint, is then created. The gradient magnitudes are sorted and added to the bin associated with its direction. Peaks in the histogram thus correspond with important directions of local gradients around the keypoint. The bin with the greatest total magnitude, and any bin with 80\% of the largest magnitude, is then used to compute the orientation of the keypoint. 

Once the keypoints location, scale and rotation are computed, the information must be converted into a descriptor vector. SIFT takes information from a 16x16 region around each  keypoint. That 16x16 region is divided into 16 4x4 arrays. Each 4x4 array is then used to calculate a 8-bin orientation histogram of the gradient magnitudes and directions of all the points in the 4x4 array.  The result is 16 8-bin orientation histograms. 

At this stage, two issues remain. The orientation histograms are dependent on rotation and illumination. To solve the first issue, we subtract the keypoint descriptor from each of the orientation histograms so that they are relative to the keypoint's orientation. To solve the second issue, we threshold each bin value to 0.2 to reduce the effects of brighter light conditions. Finally, the histograms are normalised before all 128 bin values in the 16 8-bin orientation histograms are represented as a feature vector to form the keypoint descriptor, which can later be used to match with keypoint descriptors from other images. 

\subsection{Speeded-up robust features (SURF)}

Speeded-up robust features (SURF) is a faster and more efficient version of SIFT \cite{Tareen2018}. Both methods are based around the same concept of using LoG to find the scale-space. However, SURF improves upon the LoG approximation implemented in SIFT by using box filters and integral images as opposed to the difference of Gaussians \cite{Bay2008}. SURF follows the same steps of interest point detection and keypoint feature descriptor construction \cite{Bay2008}.

\begin{figure}[H]
\centering
  \includegraphics[width=0.5\textwidth]{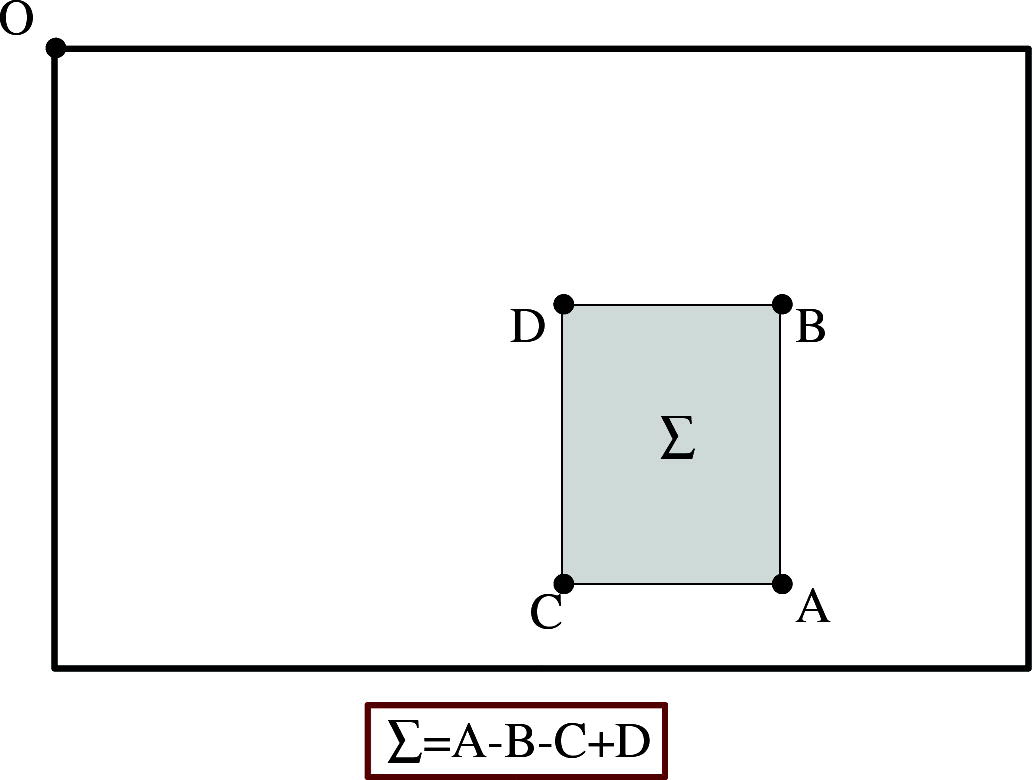}
  \caption{From \cite{Bay2008}, we can see region \(\sum = A - B - C + D\).}
  \label{fig:integral-images}
\end{figure}

Interest point detection in SURF is accomplished by Hessian matrix approximations, which are computational efficient and accurate. A Hessian matrix is a square matrix of the second order partial derivatives of a function. In this case, the function is the Gaussian function, convolved with the image \(I(x,y)\). So, given the point \(p = (x,y)\), the Hessian matrix at \(p\) at scale \(\sigma\) is:

\[
   H(p, \sigma) = 
     \begin{bmatrix}
     L_{xx}(p, \sigma) & L_{xy}(p, \sigma) \\ 
     L_{xy}(p, \sigma) & L_{yy}(p, \sigma)
     \end{bmatrix}
 \]
where \(L_{xx}(p, \sigma)\) is the convolution of the Gaussian second order derivative \(\frac{\partial^2 x}{\partial x^2}G(\sigma)\) with the image \(I\) at point \(p\), and similarly for \(L_{xy}(p, \sigma)\) and \(L_{yy}(p, \sigma)\). We then calculate the determinant of the Hessian matrix and store the results in a blob map that can be searched for interest points. The determinant of any matrix $\begin{bmatrix}
  a & b\\
  c & d
\end{bmatrix}$ is \(ac - bd\), so the determinant of the Hessian matrix is:

\[det(H) = L_{xx}(p, \sigma) L_{yy}(p, \sigma) - (L_{xy}(p, \sigma))^2\]
However, calculating the second order derivative of the Gaussian function can be computationally expensive. Instead, SURF uses box filters to approximate the second order Gaussian derivatives and leverages the concept of integral images to evaluate them at a very low computational cost. Integral images are the sum of all pixel values in a rectangular region between the origin and the location \(p = (x, y)\) in an image \(I\). With integral images it takes only four memory accesses to calculate the sum of pixel values in any rectangular region of the image, as shown in Figure \ref{fig:integral-images}. 

Thus, if we denote the box filters for the three second order Gaussian derivative calculations \(L_{xx}(p, \sigma)\), \(L_{yy}(p, \sigma)\), and \(L_{xy}(p, \sigma)\) as \(D_{xx}\), \(D_{xy}\), and \(D_{yy}\) respectively, we can find a much faster approximation for the determinant of the Hessian matrix.

\[det(H_{approx}) = D_{xx}D_{yy} - (wD_{xy})^2\]
where \(w\) is a weight applied to the filter responses to balance the expression. The Hessian determinants for all points are then calculated and stored in a blob response map. Local maxima detected from the response map are the SURF interest points. 

To create the SURF keypoint descriptor, Haar-wavelet responses in a circular region around the keypoint are calculated in both the x and y direction to find the orientation with the largest response. A 20x20 square region around the keypoint and perpendicular to the main orientation is split into 16 5x5 regions. In each region, the Haar wavelet response in the horizontal and vertical direction (with respect to the main orientation), and the sum of the absolute values of those responses, form a four dimensional feature vector. The feature vectors for the 16 sub-regions make up the a descriptor vector of length 64. 

\subsection{Oriented FAST and rotated BRIEF (ORB)}

Oriented FAST and rotated BRIEF (ORB) was developed by OpenCV as an alternative to SIFT and SURF, which are both patented and require a license to use in commercial settings (but are free for academic use) \cite{Rublee2011}. ORB borrows from two other algorithms, features from accelerated segment test (FAST), and binary robust independent elementary features (BRIEF) \cite{Rublee2011}. FAST is used for keypoint detection and BRIEF is used for creating the keypoint descriptor. 
	
FAST is a corner detection technique that examines the 16 pixels in a Bresenham circle of radius 3 around a given pixel. Using a threshold, \(t\), each surrounding pixel is checked for whether or not it is lighter, darker, or similar to the test point, \(p\). If at least 8 of those surrounding pixels are either darker or lighter than \(p\) then it is considered a keypoint. 

This process produces a high response along edges. To reduce the number of keypoints, ORB filters the keypoints using the Harris corner measure. Potential keypoints are ordered by their Harris score and the top \(N\) points are selected, where \(N\) is the number of desired keypoints.

FAST does not capture the orientation of the keypoint. ORB improves on this aspect of FAST by computing a orientation for each keypoint. The weighted intensity centroid of a patch with the keypoint at the centre is calculated and the vector from the keypoint to the centroid is the orientation assigned to the keypoint.

For creating the keypoint descriptor, ORB uses BRIEF. BRIEF constructs a binary feature vector by comparing a random pair of pixels from a neighbourhood surrounding the keypoint. The first pixel is selected from a Gaussian distribution of pixels in the image centred around the keypoint with a standard deviation of \(\sigma\). The second pixel is selected from a Gaussian distribution of pixels in the image centred around the first pixel with a standard deviation of \(\frac{\sigma}{2}\). If the first pixel is brighter than the second pixel, the value 1 is added to the binary feature vector. Otherwise, 0 is added. This process is repeated 128 times to form a 128-bit feature vector that describes the keypoint. 

Like FAST, BRIEF ignores the orientation of the keypoint. To counteract this, ORB orients the binary feature vector according the orientation detecting in the FAST stage.

For any feature set of n binary tests we have two matrices of dimensions \(2 \times n\):

\begin{align*}
    S = \begin{pmatrix}
    x_1,... x_n\\
    y_1,... y_n
  \end{pmatrix},
  S' = \begin{pmatrix}
    x_1,... x_n\\
    y_1,... y_n
  \end{pmatrix}
\end{align*}
Where \(S\) is the set of the first pixels of the \(n\) random pairs, and \(S'\) is the set of the second pixels of the \(n\) random pairs. Given the orientation of the keypoint, \(\theta\), and the corresponding rotation matrix, \(R_{\theta} = \begin{bmatrix}
  cos\theta & -sin\theta\\
  sin\theta & cos\theta
\end{bmatrix}\), we can then steer \(S\)  and \(S'\) in the direction of \(\theta\)  by doing the matrix multiplications:

\[S_{\theta} = R_{\theta}S,	S'_{\theta} = R_{\theta}S' \]
This allows ORB to build a lookup table for different values of \(\theta\) (in increments of \(2\pi/30\)), so the keypoint descriptor is more robust to image rotation. 

It is important, however, that the set of binary tests used to determine the descriptor vector is uncorrelated and has high variance. Additionally, a simple characteristic of these binary feature vectors is that their mean is close to 0.5. Maintaining a mean near 0.5 is therefore desirable, as is high variance, so that the descriptor is unique. 

To satisfy both of these goals, ORB uses a greedy search to iteratively search among all possible binary tests for 256 tests with high variance, a mean close to 0.5, and minimal correlation. This process is called rBRIEF and the result is a 256-bit binary keypoint descriptor vector. 

\subsection{Other Local Feature Extraction Methods}

Two other important feature extraction methods worth noting are local binary patterns histogram (LBPH) and histogram of oriented gradients (HOG) \cite{Ahonen2006} \cite{Dalal2005}. These are both relatively simple feature extraction techniques in comparison to SIFT, SURF, and ORB, but can still be effective.

\subsubsection{Local Binary Patterns Histogram (LBPH)}

LBPH is created using the local binary pattern operator \cite{Ahonen2006}. Firstly, the image is divided into cells. Each pixel in each cell is compared to its eight neighbouring pixels, and a new binary value 1 or 0 is assigned, depending on if the pixel's value is greater or less than the neighbouring pixel value. 0 indicates that the pixel value is greater than the neighbouring pixel value and 1 indicates 
that the pixel value is less than the neighbouring pixel value. From left to right, top to bottom, this creates a 8-digit binary number, which is converted into its decimal form. For each cell, a histogram of all the pixel values is created and concatenated to create one histogram for the entire image. The image is in grayscale for this operation, so each cell histogram has 256 bins (0-255 pixel values), and the final image feature vector is \(256 \times n\) dimensions, where \(n\) is the number of cells the image was divided into.

\subsubsection{Histogram of Oriented Gradients (HOG)}

HOG works as one might expect, by creating a histogram of the gradients within an image \cite{Dalal2005}. Like in LBPH, the image is first split into cells. Within each cell, each pixel's horizontal and vertical gradient is calculated by filtering the image with the first order derivative of the image intensity function in the horizontal direction, \(\begin{bmatrix} -1, 0, 1
\end{bmatrix}\), and the vertical direction, \(\begin{bmatrix}
 -1, 0, 1
\end{bmatrix}^\mathrm{T}\).  From the horizontal and vertical gradients, the gradient magnitude and orientation can be calculated with the following equations: 

\begin{align*}
g = \sqrt{g_{x}^2 + g_{y}^2} \\
\theta = arctan\frac{g_x}{g_y}
\end{align*}
For colour images, the gradient magnitude is the maximum of the magnitudes of the three colour channels. The orientation is the orientation of the gradient associated with the maximum gradient magnitude. 

The gradient magnitude is a value between 0 and 255, and the gradient orientation is a value between 0 and 180, because HOG uses unsigned gradients. This means we consider an angle and the angle 180 degrees opposite it the same. Unsigned gradients have been shown to be more effective at detecting humans. 

With the gradient magnitude and orientation for all pixels calculated, we can create a 9-bin histogram of gradient magnitude. Each bin represents a 20 degree bin centred at 0, 20, 40, 60, 80, 100, 120, 140 and 160 degrees. Gradient magnitudes are split proportionally between bins depending on the distance between bins. For instance, a pixel with gradient magnitude 10 and a orientation value of 130 would assign 5 to the 120 degree bin and 5 to the 140 degree bin. If the angle were 125, 7.5 would be assigned to the 120 degree bin and 2.5 to the 140 degree bin. 

Before concatenating the histogram of each cell like in LBPH, we normalise the histograms by finding the L2 norm of each histogram. This makes the vectors invariant to scale. The final image feature vector is \(9 \times n\) dimensions, where \(n\) is the number of cells the image was divided into.

\section{Feature Matching}

A significant step in the image matching process is matching. This is the process of establishing correct point correspondence between two images \cite{Ma2021}. The feature-based approach to feature matching can be split into two categories: direct matching and indirect matching \cite{Ma2021}. 

\subsection{Direct Matching}

Direct matching is a method of feature matching that aims to directly match features from two feature sets by finding the spatial relationship between two sets of features with geometric transformation and optimisation methods \cite{Ma2021}. This can be divided into graph matching and point set registration methods. 

\subsubsection{Graph Matching Methods}

Graph matching methods construct a graph assigning each feature to a node and specifying edges \cite{Ma2021}. Graphs are then matched by establishing a node to node correspondence of the feature graphs between different images. In an ideal scenario, the solution produced by GM would be a bijection of two feature graphs. That is, a one-to-one function that maps each node in graph A to a node in graph B. In real world applications, however, this is requirement is seldom possible to satisfy. 

Research in GM has focused predominantly on formulating GM as a quadratic assignment problem (QAP), which is a key problem in combinatorial optimisation \cite{Loiola2007}. Originally posed by Koopmans and Beckmann, a QAP models the real life economic problem of assigning facilities or factory plants to locations to minimise the cost of the transporting supplies between facilities \cite{Koopmans1957}. Edge weights in the graphs used to model this scenario correspond to the amount of supplies transported between two facilities. The result is that facilities which have a large edge weight and flow of supplies between them are placed closer together. In the context of feature graphs used to model feature sets, we can use the similarity between feature descriptors to determine the weight of edges between nodes \cite{Ma2021}. 

By looking at GM as a QAP, we can solve it in the same way --- by finding the one-to-one correspondence between two features set that maximises the affinity between two graphs. The affinity, J, is defined as:

\[J(X)=(K^{T}_pX) + (A_1XA_2X^T)\]
Where \(X\) is the permutation matrix between the two graphs, \(K\) is the affinity matrix between nodes, and \(A_1\) and \(A_2\) are the weighted adjacency matrices of the two graphs.

\subsubsection{Point Set Registration Methods}

Point set registration methods attempt to find the spatial transformation that optimally aligns two sets of features \cite{Ma2021}. The key difference between point set registration methods and graph matching methods is that point set registration assumes that a spatial transformation exists such that two sets of features can be overlayed with one another. By assuming more information, point set registration takes fewer parameters than graph matching and is more computationally simple. However, point set registration methods do sacrifice on robustness and generalisability.

\subsection{Indirect Matching}

Indirect matching, also known as matching with mismatch removal, is the classically favoured approach to feature matching in feature-based image matching pipelines \cite{Ma2021}. Indirect matching is composed of two stages: the construction of a preliminary match set and then the removal of poor matches by applying additional filters to the matches \cite{Ma2021}. 

\subsubsection{Preliminary Match Set Construction}

Firstly, initial correspondence between features is established by calculating the similarity between local feature descriptors. Methods for creating this preliminary set of feature matches include fixed threshold \cite{Ma2021}, nearest neighbour (NN) \cite{Brown2005}, mutual nearest neighbour (MNN) \cite{Liu2010}, and nearest neighbour distance ratio (NNDR) \cite{Lowe2004}. Nearest neighbour is also commonly referred to as brute-force matching. 

The fixed threshold strategy creates a initial match set from matches in which the distance between local image descriptors is below a set threshold \cite{Ma2021}. Fixed thresholding can perform differently in different matches because some feature set matches will naturally have lower distances among descriptors. This also means that each descriptor can have more than one matching descriptor where we would prefer that each descriptor in image A has a one-to-one correspondence with a descriptor from image B. For images that contain the same object we would expect a strong one-to-one correspondence between feature descriptors, because the same features should be identified during feature extraction (so long as the features are robust to scale, rotation and affine variations).

The NN technique is more robust than the fixed threshold strategy as it is less sensitive to specific matches, however, it also does not enforce one-to-one correspondence between feature matches \cite{Brown2005}. Nevertheless, NN detects more potentially true matches so is an improvement upon the fixed threshold strategy.

MNN is identical to the aforementioned NN technique, except that a match will only be considered a match if the descriptors are mutual best matches \cite{Liu2010}. That is, if the best match for the i-th descriptor in image A is the j-th descriptor in image B then the best match for the j-th descriptor in image B must be the i-th descriptor in image A for the match between those descriptors to be included in the preliminary match set. 

NNDR is also related to NN, however, it differs in that it considers the second nearest neighbour as well as the first \cite{Ma2021}. Matches are discarded if the ratio of the distance between feature descriptors of the first and second-best match is less than some threshold. This follows the process outlined by Lowe's (2004) ratio test, and ensures that matches are sufficiently different than the "noise", represented by the second best match, that the match provides meaningful information for matching \cite{Lowe2004}. 

\subsubsection{Mismatch Removal}

A fundamental issue with most of initial matching strategies is that they allow many matches to pass into the initial set that are not useful for matching. This is expected, however, because descriptors are naturally focused on only a small local patch of the image. For these reasons, we require a rigorous method to remove extraneous matches and retain the most important ones \cite{Ma2021}. 

The most fundamental technique for mismatch removal is random sample consensus (RANSAC) \cite{Fischler1981}. Similar to graph matching or point set registration, RANSAC assumes that two images are related by a geometric transformation function \cite{Fischler1981}. RANSAC iteratively selects a random sampling of four matches from the preliminary subset and uses those samples to project the transformation of one image onto the other. The projection is evaluated by calculating the number of other matches which lie within the projection. The transformation that yields the most inlier matches is the optimal transformation and matches which do not fit in the optimal transformation are discarded \cite{Ma2021}. 

\section{Clustering}

Clustering is perhaps the most intuitive and common unsupervised learning method.  Clustering focuses on finding natural groupings or categories within the data \cite{Saxena2017}. The goal is for instances within the same cluster to be similar to one another while instances in different clusters are dissimilar. Clustering can be used in a range of different applications and with a range of different kinds of data. Common applications of clustering include customer and market segmentation, identifying fraudulent credit card activity, recommendation engines, network analysis and search engine clustering \cite{Ghosal2020}. Clustering techniques can be broadly split into two categories: hierarchical clustering and partitional clustering \cite{Saxena2017}. 

In hierarchical clustering, instances are iteratively divided or merged until some stopping condition is met \cite{Murtagh2012}. An additional element in hierarchical clustering is the linkage method, which determines how clusters are related. In single-linkage clustering, clusters are linked by the minimum distance between any member in one cluster and any member in the other \cite{Saxena2017}. Complete linkage is the opposite, using the maximum distance between members of clusters \cite{Saxena2017}. Average or minimum variance linkage uses the average distance between any member of one cluster and any member of the other to determine the relationship between clusters \cite{Saxena2017}. 

In partitional clustering, instances are split into k-clusters, where k is set before clustering \cite{Celebi2015}. Alternative clustering techniques include grid-based, model-based, density-based clustering and spectral clustering \cite{Saxena2017}. Ultimately though, the most important factor in clustering is the similarity measure because this determines how instances are related in the cluster space. The most fundamental similarity measure is Euclidean distance, but other common measures include Manhattan distance, Minkowski distance, Pearson's correlation, and the cosine measure. 

\subsection{Image Clustering}

Clustering can also be applied to image data to segment images based on pixel intensity values for image annotation, indexing, and content-based image retrieval systems \cite{Ahmed2015}. 

A key difference between image data and other kinds of data is that image data contains spatial data in the form of x and y coordinates in addition to the pixel feature vector \cite{Tran2005}. In other data we might not expect the spatial relation of features in the feature vector to have importance, but spatial data is crucial for image data. 

However, there are issues with clustering directly with image feature vectors \cite{Tran2005}. Firstly, the size of the image is important because the larger the image the larger the feature vector. For image data the total number of pixels in a dataset can be larger than a million pixels. This can be prohibitive in terms of memory and processing time for clustering techniques such as hierarchical clustering, which constructs a pairwise distance matrix for all instances. 

Image clustering is therefore much more efficient when we first extract features to construct a feature vector for each image and cluster images using their corresponding feature vectors.

\section{Related Work}

Recent studies in individual species identification have typically taken a supervised approach \cite{Ferreira2020}. Ferreira et al. investigated deep learning-based methods for individual recognition in small birds \cite{Ferreira2020}. A convolutional neural network (CNN) was trained with image data collected from cameras using radio frequency identification (RFID) that auto-labelled images according to the bird’s passive integrated transponder (PIT) tag. The model performed well on birds in the training set, scoring 87.0\% for zebra finches, 90\% for great tits and 92.4\% for sociable weavers. They admitted a limitation of the model was handling birds not exposed to the model during training but did propose a method of thresholding the entropy of the softmax probability distributions for each individual to determine if the bird was likely to be a new individual. Yet, there was still significant overlap between entropies of known and unknown birds such that at least one in six unknown individuals would be misclassified as a known bird. This study highlights the impracticality of a supervised approach to individual identification. Despite high classification accuracy for all three species, the model is limited to birds learned during training. Given this constraint, and that all birds that the model was trained on are already fitted with a PIT tag, it begs the question of when their automated image classification model would be used instead of RFID. 
	
Another study focusing on identifying individual giraffes also takes a supervised approach, however, it wraps an unsupervised approach within it \cite{Miele2021}. Rather than training the data on labelled data, Miele et al. created a re-identification pipeline \cite{Miele2021}. Firstly, images are cropped using CNN-based object detection. Scale-invariant feature transform (SIFT) features are then detected and superimposed with a geometric transformation called homography to find the most relevant matches. The Euclidean distances between those relevant pairs of keypoints were measured to give a final sift-based distance measure. Finally, the sift-based distances were used to create a similarity network, that uses a property of complex networks called explosive percolation to create clusters. Explosive percolation is the threshold at which, when the distance between nodes is increased slightly, all nodes are connected. By experimenting with different threshold values and evaluating the results graphically, a threshold value was selected where each cluster theoretically represented an individual giraffe. This concludes the unsupervised section of their process. The process becomes supervised when a CNN is then trained using the images and their associated cluster label, essentially treating the cluster label as the true identity. Similar to the previous study by Ferreira et al., Miele et al. reported an impressive Top-1 accuracy of \textgreater 90\% and a false-positive rate of \textless 1\% \cite{Ferreira2020} \cite{Miele2021}. Such a low false-positive rate is impressive because it means that it was almost always correct when an individual was predicted. Nevertheless, the model is still restricted to predicting known individuals and is unable to grow the label set. Additionally, there is a significant trade-off between the true-positive rate of known individuals and the true-negative in predicting unknown individuals. The optimal balance between the two results in approximately 80\% true positive rate in known individuals and a true negative rate in unknown individuals. 

Of course, the specific requirements of each individual intraspecies classification system are different. The giraffe population of the Hwange National Park, Zimbabwe, is not growing as quickly as the kākā population of Zealandia. Retraining a model with new images once per year is a more viable option in the case of giraffes and being able to constantly update is not as important for a model to be practically useful for ecologists and conservationists in Hwange National Park. This is not the case for the kākā. While the supervised section of the work presented by Miele et al. again illustrates its drawbacks in the circumstance of population flux, the unsupervised identification pipeline provides a model for the kind of system that we aim to implement in this research. 

\section{Dataset}\label{sec:dataset}

\begin{figure}[H]
\centering
  \includegraphics[width=0.8\textwidth]{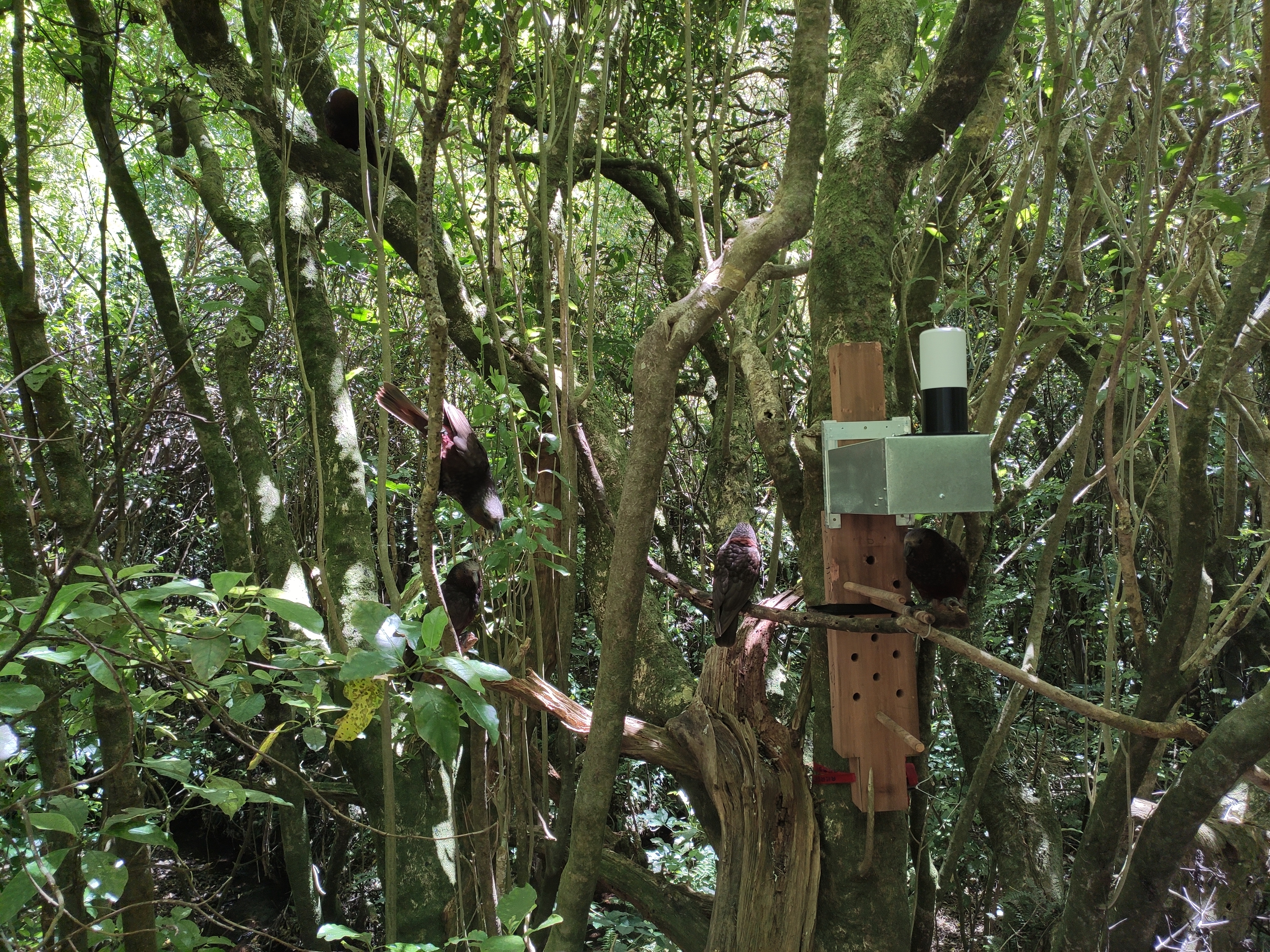}
  \caption{The kākā feeder during an active period of kākā feeding.}
  \label{fig:kaka-feeder}
\end{figure}

The dataset used in this research was created from video footage of kākā at the kākā feeding site in Zealandia Te Māra a Tāne, the Wellington wildlife sanctuary, as shown in \ref{fig:kaka-feeder}. The footage was collected during my summer research project, between November 1 and December 1, 2021. At the feeding site, a feeding station equipped with a GoPro camera with motion detection recorded footage of the kākā as they were in the motion of feeding. 

I set up the feeding station in the morning or late morning approximately four days of the week and record footage until the battery of the GoPro ran out. In a typical day, this was between three and five hours of footage, depending on the amount of kākā activity. I also visually monitored the feeder for the majority (approximately 90\%) of that time. If the kākā that visited the feeder were banded, the time and colour of the kākā's band combination were noted to be corroborated later with the timestamp of the footage recorded by the GoPro and used to label some of the dataset. Labelled images were labelled according to the kākā's coloured band combination.  

Care was taken to position the nozzle of the feeder and the ledges beneath the feeder such that the kākā presented the profile of their beak to the camera. Additionally, a matte white plastic cover was used so that only the tip of nozzle is visible in the background of the footage. 

Images were extracted from the video footage by selecting frames from the video footage that contained fully visible kākā feeding with the full profile of their beak visible. This was accomplished by a relatively naive approach where images were extracted when the pixel intensity of a point at four fifths of the height of the image and half of the width of the image is above a threshold of 50.

The video footage dataset contains 1462 video clips of varying duration. Some clips are as short as a three seconds, some are as long as a minute, although the majority are between five and fifteen seconds long. From those 1462 video clips, 6778 images with a resolution of 1920x1080 were extracted. After pre-processing, 803 images were removed, leaving 5975 images. Those images were cropped to the dimensions 540x768 to remove as much of the background feeder cover without removing relevant part of the kākā.

\begin{figure}[H]
  \centering
  \begin{minipage}[t]{0.55\textwidth}
    \includegraphics[width=\textwidth]{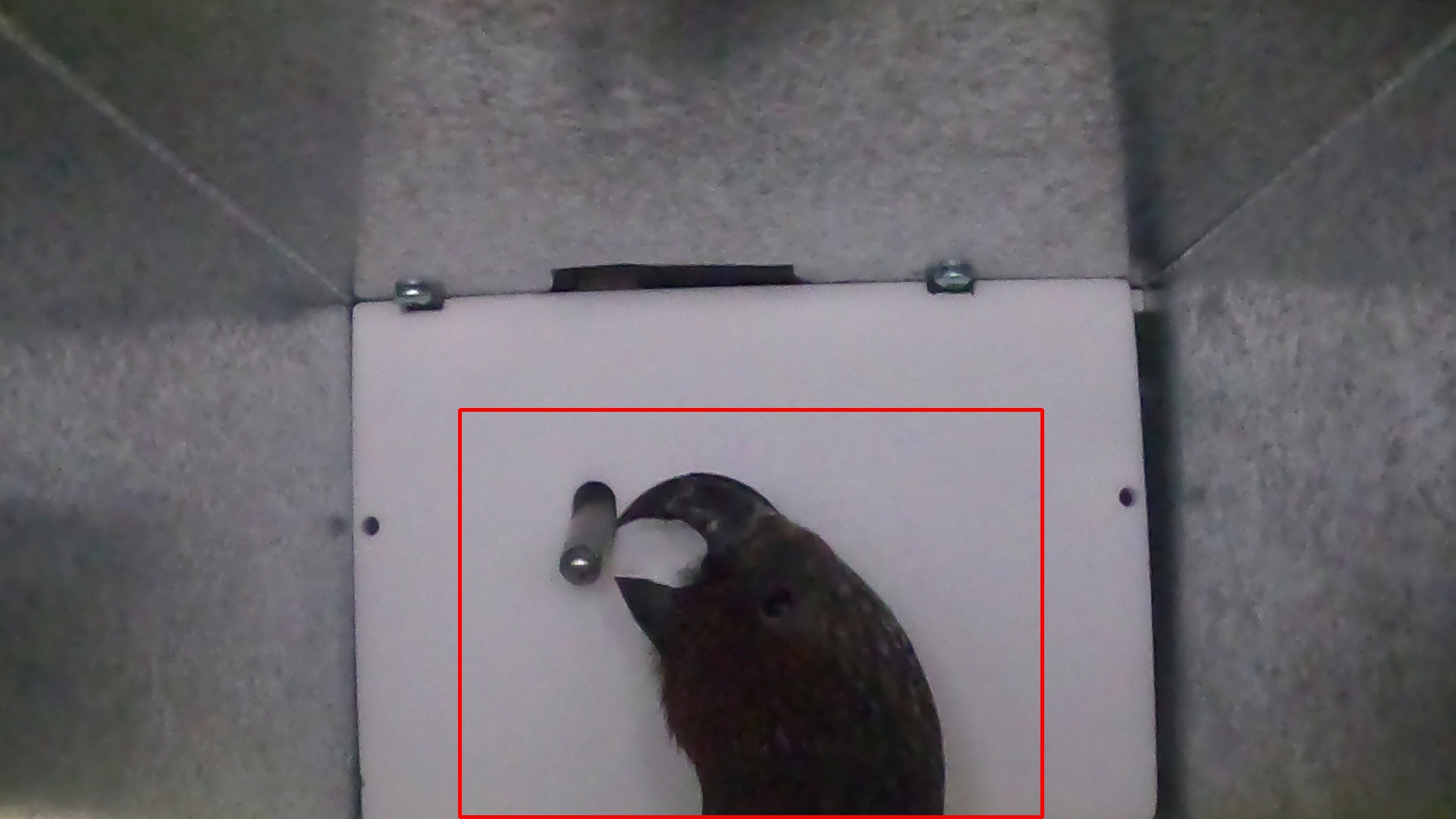}
    \caption{Image of kākā extracted from the video footage.}
  \end{minipage}
  \hfill
  \begin{minipage}[t]{0.44\textwidth}
    \includegraphics[width=\textwidth]{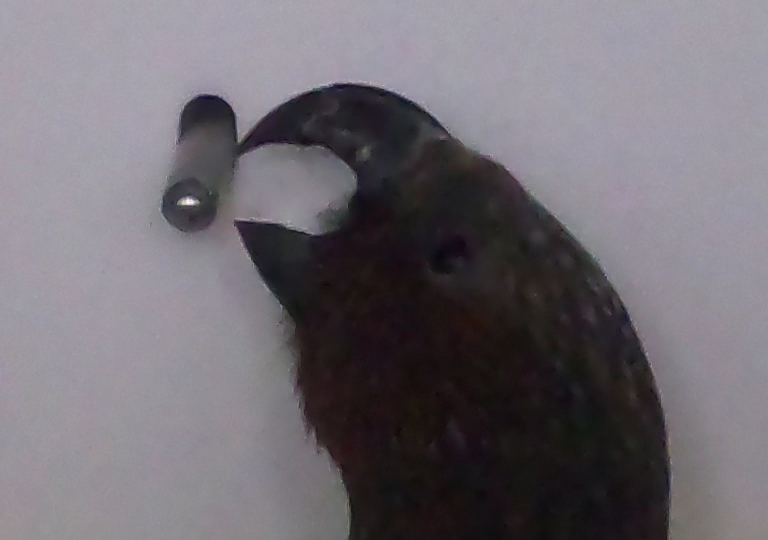}
    \caption{Image of kākā extracted from the video footage after preprocessing.}
  \end{minipage}
\end{figure}

Of those 5975 images, 795 are labeled one of seven labels: Lime-PurpleBlue, LimePurple-Green, Orange-RedSilver, PurpleRed-Red, WhiteSilver-Pink, Yellow-Green-Purple and YellowPurple-Yellow, according to their coloured band combination.

\chapter{Methodology}\label{C:methodology}

\section{Initial Method}

The goal of this research is to develop a image matching pipeline that would be used to inform a similarity network and cluster images by individuals. Work on this project will therefore follow the image matching pipeline and iteratively improve each stage of the process until sufficient results are produced as to continue to the next stage of the pipeline.

The process initially began with exploring feature extraction methods to discover image features and gain an understanding of the capabilities and potential of a feature-based approach to image matching. However, after initial feature extraction and feature matching it became clear that object localisation was an essential component of the image matching pipeline. 

\section{Proposed Method}

The main idea of the proposed method is to follow a image matching pipeline to obtain a similarity measure for comparing images and constructing an online and unsupervised image similarity network to discover clusters in. 

We will use a feature-based image matching pipeline because it is more flexible and robust than area-based image matching and and more established than deep-learning approaches to image matching, which are still being developed. Important stages in this process are object localisation, feature extraction, feature matching and image matching.

\subsection{Object Localisation}

As discussed, object detection to localise the kākā in each image is essential for extracting important features that are relevant to the kākā and will be accomplished by k-means image segmentation. K-means image segmentation is a very simple segmentation method, however, it can be effective for object detection in images where the background is a consistent colour and without noise, such as the images in the dataset used in this research.

\subsection{Feature Extraction Method}

For the feature-based image matching process, SIFT was the chosen feature extraction method. SIFT is an established and well-known feature extraction algorithm that has proven its ability to achieve good results in image matching. 

The key measure used to choose the feature extractor was accuracy, while speed and efficiency were secondary concerns. With an online approach, features will only be extracted from an image once, when that image is added to the model. Therefore, so long as the speed and efficiency of the extractor are not entirely prohibitive, accuracy is paramount.

In this respect, SIFT performs the best out of SIFT, SURF, and ORB across variations in scale, rotation and affine variations \cite{Karami2015}. This was supported by another study, which compared a wide range of local feature extractors including SIFT, SURF, ORB, and other extractors such as KAZE, AKAZE and BRISK to conclude that SIFT was the most accurate of them all \cite{Tareen2018}. While SURF and ORB are both viable alternatives, they excel in efficiency and speed \cite{Karami2015}. As mentioned though, this is not the priority of the feature extraction method used in this research. 

Finally, SIFT being so well-supported by existing Python computer vision packages makes it simple to implement and use. This will be further discussed in Section \ref{sec:algorithm-implementations}.

\subsection{Feature Matching}\label{sec:proposed-feature-matching}

For feature matching we will qualitatively compare the NN, MNN and NNDR matching strategies, in combination with RANSAC mismatch removal. Indirect matching methods were chosen because they have been more classically favoured than direct matching methods, and these three methods were chosen in specific because they are more robust than the fixed threshold method. 

\subsection{Image Matching}

For image matching all features in each image are compared with the features of all other images in the dataset, except for images from the video clip that the image being compared with all others was extracted from. Images extracted from the same video clip as the image being matched with are not considered by the image matching function, although images extracted from adjacent video clips are (i.e. video clips taken on the same day, before or after).

\subsubsection{Similarity Measure}\label{sec:similarity-measure}

The similarity measure is used to calculate the best matches for each image in the dataset, and is calculated using the number of matches between feature descriptors from separate images and the average distance between those descriptors. 

The number of matches between feature descriptors is the most robust measure for the quality of a match because matches must pass through preliminary matching methods and rigorous mismatch removal. The average distance between descriptors is also a useful measure of match quality, but it is somewhat incidental to matches. Naturally, feature descriptors that have a small distance between them are more likely to match, but one match with a larger distance between its matching descriptor may drastically increase the overall average distance between descriptors, particularly when there are not many feature matches.

Consequently, the similarity measure should give more weight to the number of matches than average distance, however, it should still incorporate average distance to discriminate between image matches which have the same number of feature matches. We define the similarity \(S\)  between two images as follows:

\[ S(D) = |D| + \frac{1}{1 + \sum_{n=1}^{D} \frac{d_n}{|D|}}\]Where \(D\) is the set of distances between matches, \(d_n\) is the n-th distance in the set \(D\), and \(|D|\) is the number of total matches. 

By formulating similarity like this, an image match which has more features matches than another image match will always be ranked higher. Furthermore, image matches that have a low average distance between feature matches will rank higher than other image matches with the same total number of feature matches. The minimum value of the match term on the left of the addition is four, because RANSAC require a minimum of four sample matches to project transformations with. The maximum value of the average distance term on the right of the addition is one. This would be achieved by matching an image with itself, because the distance between all descriptors would be zero. 

\section{Experiment Implementation}

The proposed method has been implemented using OpenCV, an open source computer vision and machine learning software library. 

The code used in this research uses OpenCV version 4.3.0, specifically, because this version allows the use of SIFT by default. From OpenCV 3 onwards, SIFT was excluded from OpenCV from their default library and included in their non-free algorithms set because the SIFT detector is patented by the University of British Columbia. This patent expired in March 2020, so OpenCV included SIFT in their default library in the next version of OpenCV, 4.3.0. 

\subsection{Algorithm Implementations}\label{sec:algorithm-implementations}

OpenCV provides implementations of SIFT feature detection and description as well as implementations of common feature matching and mismatch removal algorithms.

With OpenCV, SIFT feature detection and description is incredibly simple and can be executed with one line of code. 

As mentioned, OpenCV also provides implementations of common feature matching algorithms. For this task, OpenCV has both a brute force (BF) matcher and a fast library for approximate nearest neighbours (FLANN) matcher. Both of these are implementations of the NN feature matching algorithm. The BF matcher will exhaustively match every descriptor from two images, finding the best possible match for each descriptor, whereas the FLANN matcher will find the approximate nearest neighbours. This is much more efficient but the matches are not necessarily optimal.  For finding feature matches within small datasets, the accuracy of the BF matcher negates the slower processing time, but for much larger datasets, the efficiency of the FLANN matcher would be worth considering if the cost of potentially reducing accuracy was worth increasing the efficiency. For the parameters of this research, the BF matcher is preferred because it is more accurate. Additionally, efficiency is not as important because an online similarity network using the image matching pipeline would only calculate the feature matches once.

OpenCV also provides for implementing MNN and NNDR matching strategies. 

MNN can be implemented by setting the second parameter of the BF matcher object, \textit{crossCheck}, to true. This parameter is a boolean variable which determines whether descriptors must be mutual best matches to be considered a match. This enforces a one-to-one correspondence among matches as per the MNN matching strategy. 

NNDR can be implemented by using the \textit{knnMatch} method to find matches between descriptors rather than the default \textit{match} method.  The \textit{knnMatch} function takes in a parameter \(k\) and returns the \(k\) best matches, in order. By running \textit{knnMatch} with \(k = 2\), we can find the top two matches for each descriptor and perform the ratio test on the distances between the top two descriptor matches, discarding matches that are not sufficiently different. 

\section{Evaluation}

To evaluate the performance of the image matching function we calculate the best image match for each image from the labelled subset of the dataset and then check that the best match for each image shares the same label. This produces an accuracy score which gives an approximation of how well the image matching function performs.

\subsection{Labelled Subset}\label{sec:labelled-subset}

The data used to test the overall functionality and accuracy of the image matching pipeline is the labelled subset of the overall dataset. These labels were collected during the data collection stage as outlined in Section \ref{sec:dataset}. 

\begin{table}[H]
\caption{Labelled Subset of the Data Used for Evaluation} 
\centering 
\begin{tabular}{c c} 
\hline 
\textbf{Label} & \textbf{Images} \\ 
\hline
Lime-PurpleBlue & 139\\ [1ex]
LimePurple-Green & 102 \\[1ex]
Orange-RedSilver & 202\\[1ex]
PurpleRed-Red & 57\\ [1ex]
WhiteSilver-Pink & 139\\ [1ex]
Yellow-GreenPurple & 9 \\ [1ex]
YellowPurple-Yellow & 147\\ [1ex]
\hline 
 Total: &  795\\
\hline 
\end{tabular}
\label{table:labelled-subset}
\end{table}

The split of labels in the labelled subset of the data is shown in Table \ref{table:labelled-subset}. The wide variance in the number of images for each label gives a way to evaluate the specificity of the image matching method. If the image matching function is able to match images from a label set containing only 9 images, this indicates that the image matching pipeline would be capable of detecting new individuals with very few images. For most of the labels though, there are between 100 and 200 images for each label because this would be in the range of the expected number of images for any given individual in the dataset, depending on how frequently the kākā visited the feeder during data collection. 

In supervised learning, a test set in machine learning would be a set of instances unseen by the model. However, in unsupervised learning it is more difficult to evaluate predictions because there are sometimes no ground truth labels. However, datasets used in unsupervised learning are not necessarily unlabelled, they are just not used as part of the learning process. We can still use labelled data to evaluate the performance of the unsupervised learning method. In this way, we use the labelled subset of our dataset to measure how effective the image matching pipeline is at producing accurate image matches.

Note that the images in the evaluation set were labelled as the data was collected, but not all instances of kākā that had a band combination feeding were observed, so there may be images in the dataset that are unlabelled but contain the same kākā as images that are labelled. If one of these images is matched with by a labelled image containing the same kākā, the match will be falsely classified as an incorrect match (but correctly classified as sharing the same label). Therefore, the accuracy measure is likely an underestimate of the accuracy of the image matching function.

\chapter{Image Matching Pipeline}\label{C:imp} 

\section{Object Localisation}\label{sec:object-localisation}

Based on initial testing, we concluded that object localisation was necessary to produce more relevant features during feature extraction. The main problem observed in the initial feature extraction results was that the nozzle of the feeder generates a high number of feature matches between images. These features are heavily favoured by feature matching strategies because the reflective metal of the nozzle produces high contrast among pixel values. Furthermore, the nozzle is always in-focus and highly detailed in images from the dataset because the nozzle is stationary across video frames. 

Because the nozzle is present in all images in the dataset, these features could theoretically be ignored because they do not give an advantage to any one image in the matching process. However, the nozzle is visible to varying degrees in each image, depending on if it is obscured by the kākā's head or not. Consequently, images which have the same view of the nozzle will be disproportionately matched together despite differences in the kākā contained in the image. 

Additionally, the goal of this research is match images based upon similarities between the kākā in the images. With a feature-based matching approach this means that features should be located on the beak of the kākā because the beak is where the most visually distinct features of each kākā are located. Of course, there are differences between the birds in their plumage, arrangement of feathers and geometry of their beak. However, these features (aside from geometry) are susceptible to change over time whereas markings on kākā’s beaks are high-quality and unique features throughout the kākā’s lifespan. Particularly as this image matching pipeline is intended to be used for constructing an online similarity network, its important that features are of a high quality because learning low-quality and temporary features might cause the model to require retraining. 

Based on feedback on preliminary research, we first investigated image background removal techniques that might remove the nozzle from the images. This technique uses motion detection to detect the difference between the background and the foreground in a clip of video footage. This would directly produce a mask for the foreground containing the kākā, that could be used to exclude keypoints that were not located on the kākā. However, this would represent a large divergence from the research done thus far. Images had already been extracted from the video footage and returning to the video footage format of data would introduce the difficulty of finding the frames of images in the dataset from the original video footage and extracting the corresponding foreground mask for each image in the dataset. 

Applying object detection to video footage also seemed an excessive use of resources and time given that the background is most unchanged between images. The only minor difference is that the exact location of the nozzle in the frame slightly differs between images. These differences could be due to the kākā interfering with the camera during the collection of footage or simply variance in the angle and position of the camera as it was manually setup for footage each day.

The initially approach taken was a naive and simplified method. Rather than using motion detection to find a mask for the foreground, we can extract a mask of the foreground by superimposing a mask of the background with the foreground mask of any target image to remove the nozzle. 

We therefore took a more simplified approach to object detection, using k-means image segmentation to segment the background from the foreground. This process step alone will not remove the nozzle. Image segmentation is not always generally applicable to object detection tasks, but the background of the images in the dataset is a consistent and uniform colour with relatively little noise, so image segmentation is therefore a viable solution.

\begin{figure}[H]
  \centering
  \begin{minipage}[t]{0.48\textwidth}
    \includegraphics[width=\textwidth]{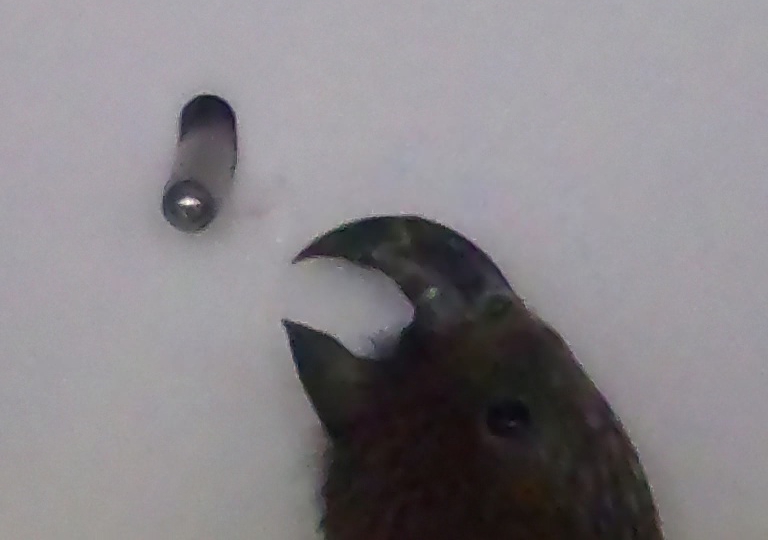}
    \caption{Image of kākā before k-means image segmentation.}
     \label{fig:before-kmeans}
  \end{minipage}\hfill
  \begin{minipage}[t]{0.48\textwidth}
    \includegraphics[width=\textwidth]{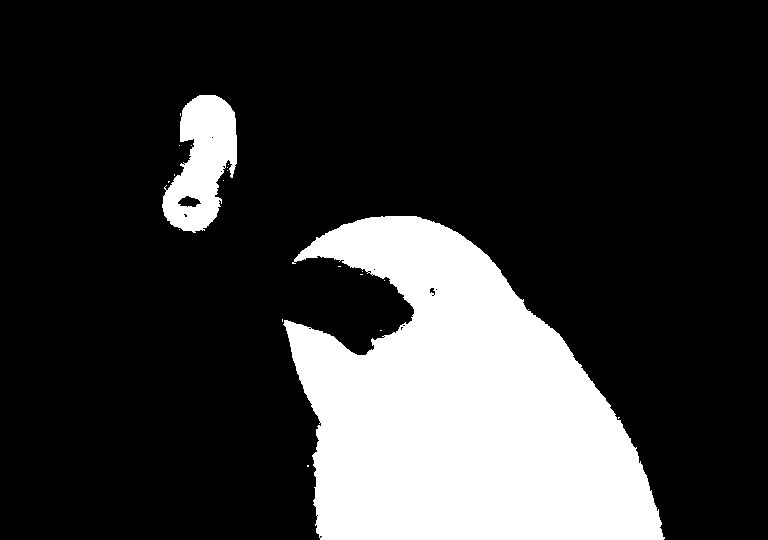}
    \caption{Image of kākā segmented with n clusters = 2.}
    \label{fig:kaka-2-clusters}
  \end{minipage}
\end{figure}

We perform k-means image segmentation with \(k = 2\), with the before and after results shown in Figure \ref{fig:before-kmeans} and Figure \ref{fig:kaka-2-clusters} respectively.

\begin{figure}[H]
\centering
  \includegraphics[width=0.6\textwidth]{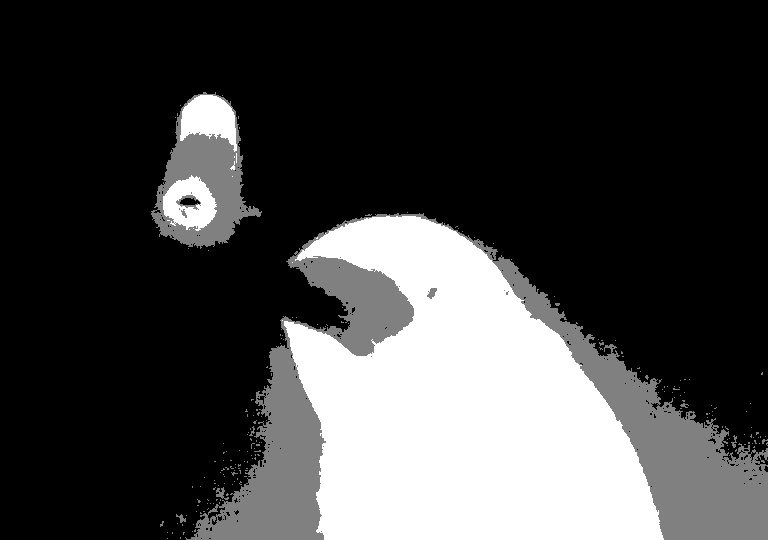}
    \caption{Image of kākā segmented with n clusters = 3.}
    \label{fig:kaka-3-clusters}
\end{figure}

As we can see, segmenting the background from the nozzle is a simple process process, and visually this makes sense. If we squint our eyes at the image of the kākā shown in  Figure \ref{fig:before-kmeans} we can extract a similar segmentation ourselves. However, it is more complicated to segment the nozzle from the kākā. If we cluster the image into three clusters like, meaning k-means segmentation with \(k = 3\), we can see in Figure \ref{fig:kaka-3-clusters} that increasing the value of \(k\) is ineffective. 

Rather than segmenting the image into three continuous regions or objects like we might wish, it simply identifies regions where the colour intensity is between the darker pixel intensity of the kākā and parts of the nozzle, and the light pixel intensity of the background. Of course, this a good example of the limitations of k-means image segmentation in object detection tasks because it works by segmenting based on raw pixel values. 

We proceed by applying some of the principles of image background subtraction using the masks produced by k-means image segmentation. If we calculate a mask of the background without the kākā present we can subtract the background mask from a mask of the foreground containing the kākā, such that shown in Figure \ref{fig:kaka-2-clusters}. 

\begin{figure}[H]
  \centering
  \begin{minipage}[t]{0.48\textwidth}
     \includegraphics[width=\textwidth]{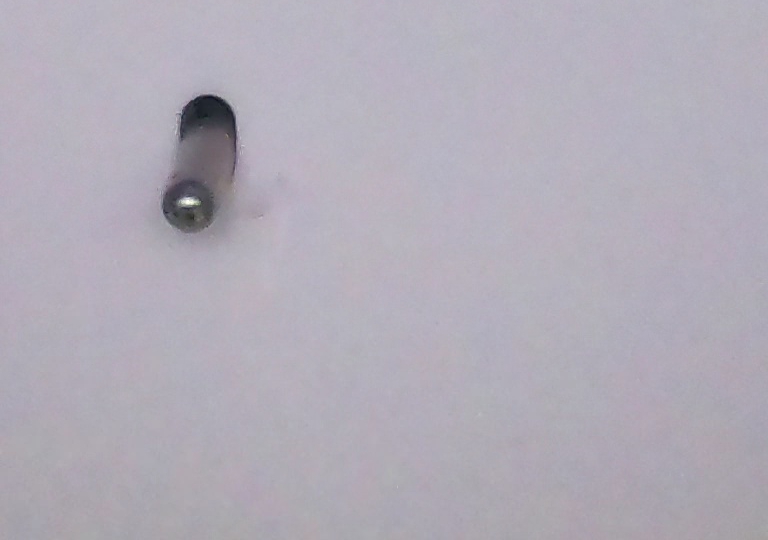}
     \caption{Sample background image}
     \label{fig:bg}
  \end{minipage} \hfill
  \begin{minipage}[t]{0.48\textwidth}
    \includegraphics[width=\textwidth]{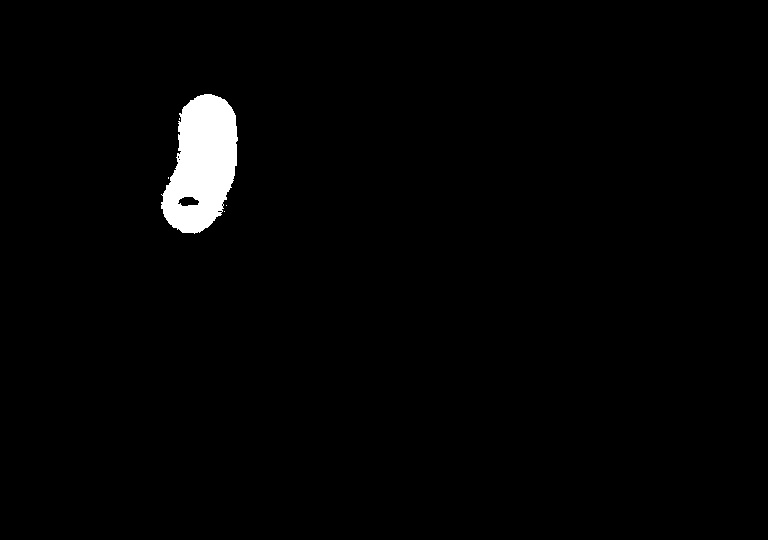}
  \caption{Mask of the sample background image}
    \label{fig:bg-mask}
  \end{minipage}
\end{figure}

For initial tests, we used a sample background image, shown in Figure \ref{fig:bg}, taken from the last frames of the first clip of footage in the dataset to reduce complexity. However, using a sample image does introduce problems when the positioning of the nozzle slightly differs between the sample and true background for an image which we will discuss solutions to. This step could be improved upon by extracting a background image from each clip of video footage and constructing a secondary dataset with the background specific to each video clip. 

Using this sample image, we create a mask separating the nozzle from the rest of the background with k-means image segmentation using \(k = 2\), shown in Figure \ref{fig:bg-mask}.

It's important to note that this implementation of k-means image segmentation uses random centroid initialisation. So, depending on where the centroids are randomly initialised, the binary number value assigned to the background varies. For the mathematics of creating the object localisation mask it is important that values assigned to specific regions are consistent. To normalise the background and foreground masks across different runs of k-means segmentation, we assign the value 1 to regions of the mask that represent desirable matches and the value 0 to undesirable matches. For the background mask, the nozzle region is therefore composed of 0s while the rest of the mask contains 1s. For the foreground mask containing the kākā and the nozzle, pixels in the kākā and nozzle regions are assigned the value 1. To create the kākā localisation mask to filter keypoints we then superimpose the background and the foreground mask and only consider keypoints where the the sum value of the background and target mask at that keypoint is 2.

\begin{figure}[H]
\centering
  \includegraphics[width=0.6\textwidth]{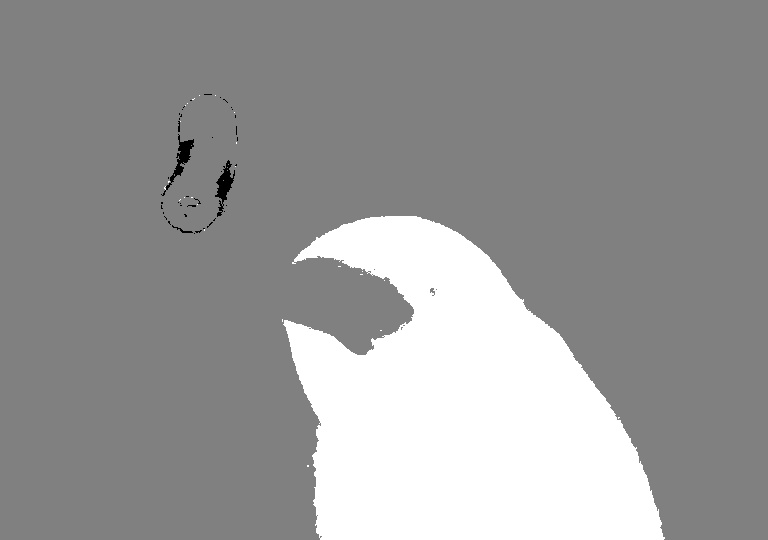}
  \caption{Example mask after superimposing the background mask with the foreground mask.}
  \label{fig:bg-subtracted}
\end{figure}

As shown in Figure \ref{fig:bg-subtracted}, this method works well when the image has the exact same background as the background mask. The white regions of \ref{fig:bg-subtracted} indicate where the value of the kākā localisation mask is 2, the grey regions indicate a value of 1, and black regions indicate a value of 0.  As mentioned, however, there is slight variance between images in the location of the nozzle as well as the lighting of the background. Consequently, superimposing the background only results in a clean removal of the nozzle for images with the exact same positioning of the nozzle. For other images, the nozzle is still partially visible because the nozzles do not directly overlay one another. 

Regardless, however, there is still noise present because k-means segmentation is not deterministic. This means that even though the background mask and the foreground mask used to calculate the localisation mask shown in Figure \ref{fig:bg-subtracted} are from the same piece of video footage, the segmentation around the nozzle is minutely different. We can see this in Figure \ref{fig:bg-subtracted} where the white pixels scattered around the black outline of the nozzle represent pixels that have "slipped" through mask superimposition and may allow keypoints that are not on the kākā to be used for comparison.

To solve this, we use a 9x9 convolution filter to blur the mask. This removes any parts of the nozzle that may register as still visible after being filtered through the background and target image mask. 

\begin{figure}[H]
\centering
  \includegraphics[width=0.6\textwidth]{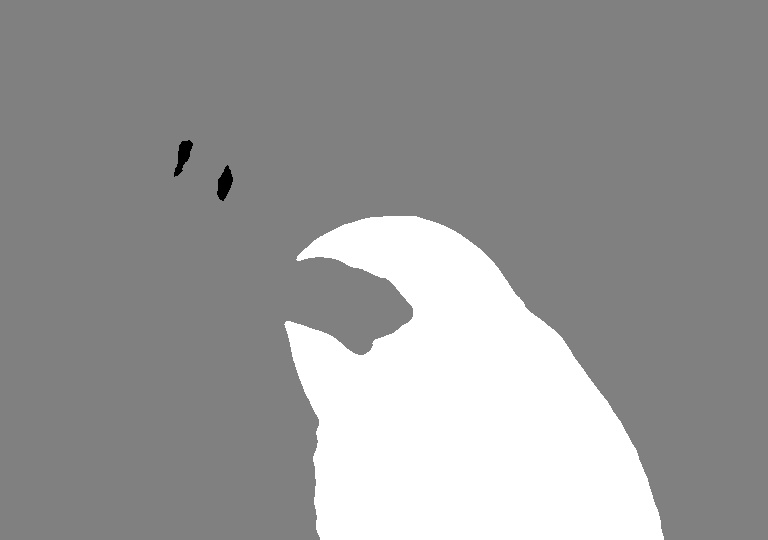}
  \caption{Example mask after superimposing the background mask with the foreground mask and blurring.}
  \label{fig:bg-subtracted-blurred}
\end{figure}

\begin{figure}[H]
\centering
  \includegraphics[width=0.6\textwidth]{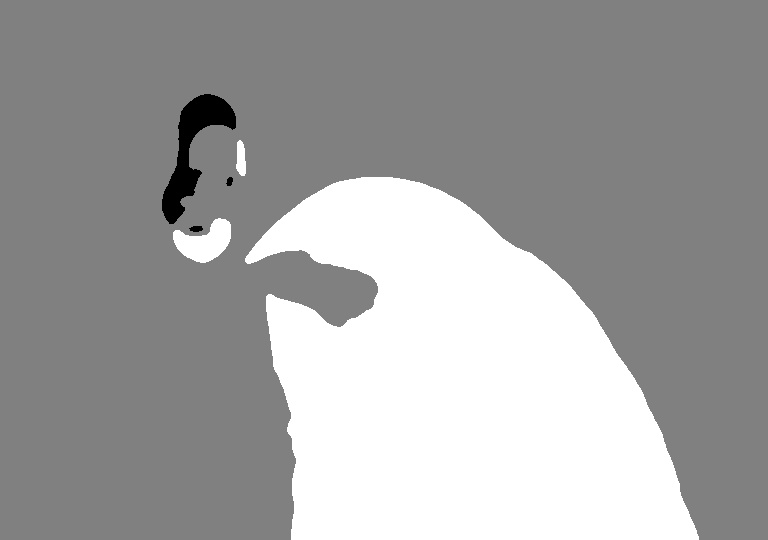}
  \caption{Example of a partially visible nozzle mask even after mask blurring.}
  \label{fig:bad-ex-bg-subtraction}
\end{figure}

As mentioned, however, there are still instances like those shown in Figure \ref{fig:bad-ex-bg-subtraction} where the nozzle is still partially visible in the localisation mask after blurring. 

\begin{figure}[H]
\centering
  \includegraphics[width=0.6\textwidth]{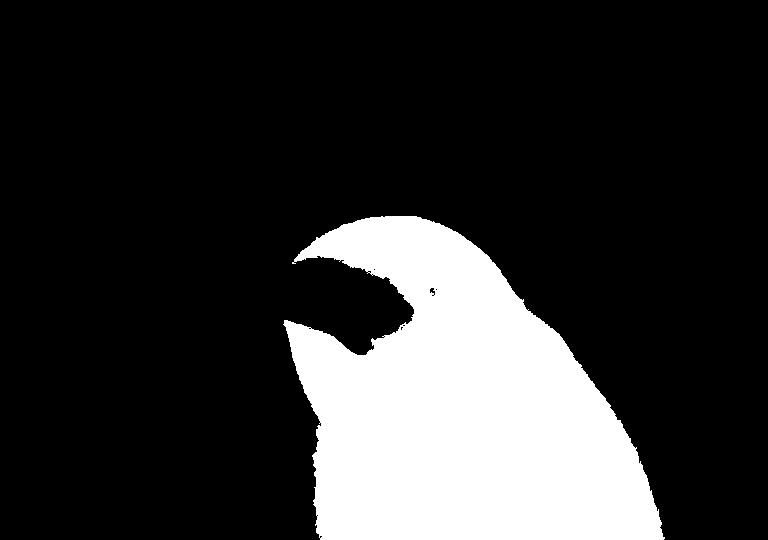}
  \caption{Foreground mask with the nozzle removed.}
  \label{fig:blobs-removed}
\end{figure}

To solve this issue, we can exploit a simple property of the localisation mask, which describes the fundamental difference between the nozzle and the kākā: the number of pixels within each object. From the initial target mask that is created by k-means segmentation we can remove "blobs" or clusters of pixels of the same value that contain fewer than some threshold such that the nozzle is removed but the kākā remains. Now, rather than relying on the nozzles sufficiently overlapping between the background and foreground mask, we can consistently remove the nozzle from the foreground mask, given that the nozzle "blob" and kākā "blob" are separate from one another. The result of this is shown in Figure \ref{fig:blobs-removed}.

Still, however, this method is not without flaws. As shown in \textit{Figure 4.16}, when the kākā's beak and the nozzle overlap one another, the nozzle is not removed. This is because they form one continuous "blob" so there is no smaller "blob" to remove via thresholding.

\begin{figure}[H]
\centering
  \includegraphics[width=0.6\textwidth]{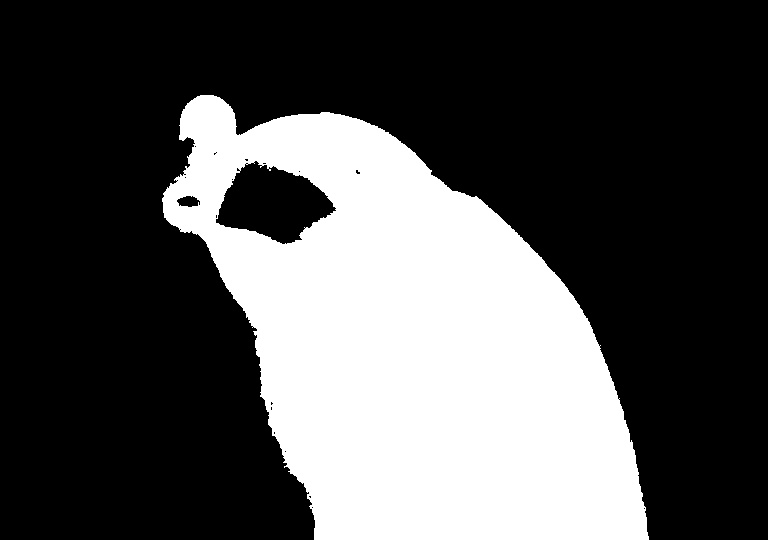}
  \caption{Example clustering where the kākā and nozzle are part of the same "blob".}
\end{figure}

\begin{figure}[H]
\centering
  \includegraphics[width=0.6\textwidth]{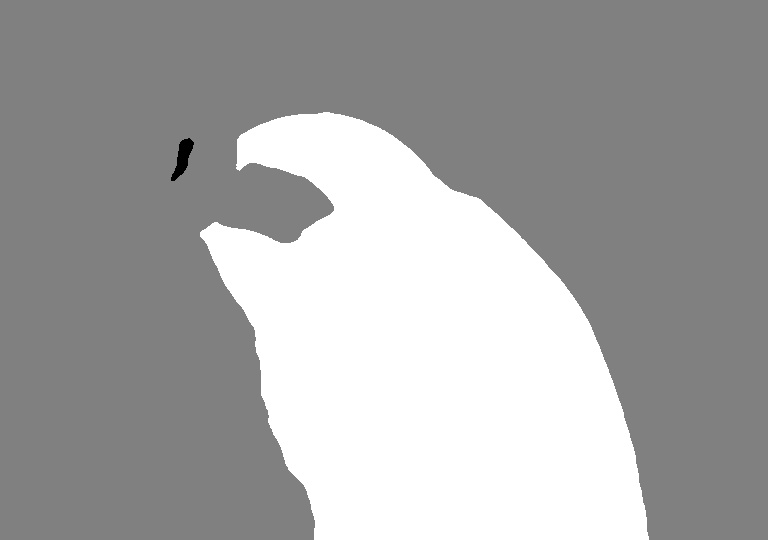}
  \caption{Mask of an example image after the nozzle has been removed by superimposing a background mask.}
  \label{fig:blob-fail}
\end{figure}

For masks such as that shown in Figure \ref{fig:blobs-removed} we can follow the original process of creating a localisation mask by superimposing the background mask and the foreground mask. The results of this are shown in Figure \ref{fig:blob-fail}. The nozzle is successfully removed, but we can see that the features at the tips of the upper and lower mandible of the kākā beak would be passed over after being filtered through the localisation mask because the beak overlaps with the nozzle. When it is just the tips of the beak that are cut short, we consider this acceptable because there are relatively few features detected in that region of the beak. Most of the important keypoints are detected further up the beak. 

\begin{figure}[H]
\centering
  \includegraphics[width=0.6\textwidth]{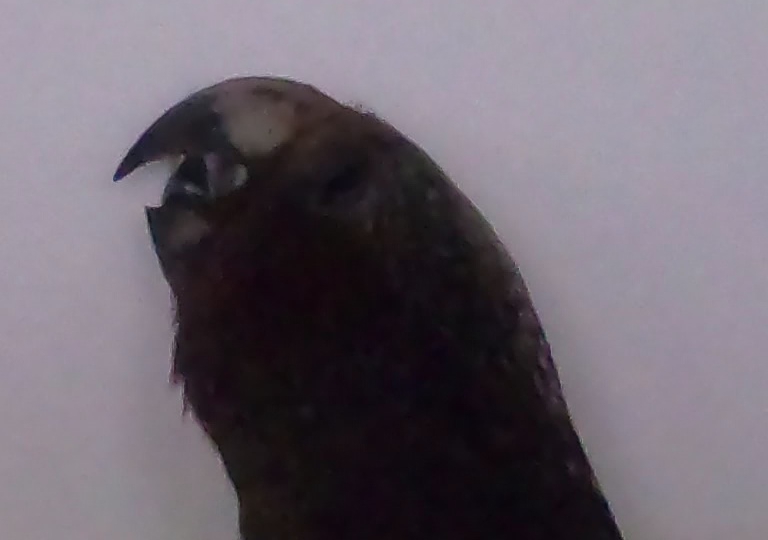}
  \caption{Example of an image where the kākā is obscuring the nozzle.}
  \label{fig:kaka-obscuring-nozzle}
\end{figure}

However, some of the kākā's feeding behaviour led to their beak obscuring most of the nozzle, as shown in Figure \ref{fig:kaka-obscuring-nozzle}.

\begin{figure}[H]
\centering
  \includegraphics[width=0.6\textwidth]{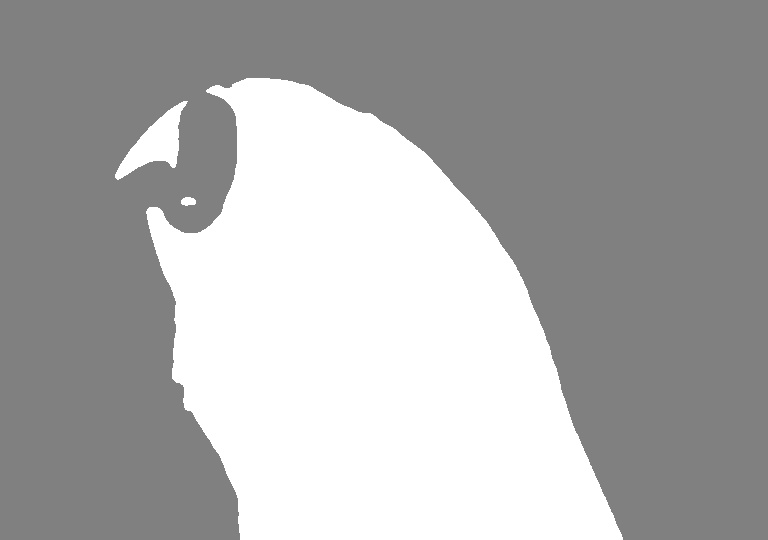}
  \caption{Mask of an example image where the kākā is obscuring the nozzle.}
  \label{fig:blob-fail2}
\end{figure}

The mask for the image shown in Figure \ref{fig:kaka-obscuring-nozzle} would be as pictured in Figure \ref{fig:blob-fail2} when following the object localisation process defined thus far.

In this situation, the localisation mask would ideally filter only the non-obscured sections of the nozzle. Yet, this mask would not remove all of the features along the beak and the features that remain after being filtered with the localisation mask in Figure \ref{fig:blob-fail2} would still be useful for feature matching and image matching. Albeit, the effectiveness of feature match would increase if the missing relevant features could be recovered. Most importantly though, it would remove the features that would be detected around the tip of nozzle and entirely change feature matching outcomes. 

Overall, the mask shown in Figure \ref{fig:blob-fail2} illustrates one of the largest flaws with our approach to localising the kākā as well as with data collection in the "real world". 

During data collection we anticipated removing the nozzle to be a large problem, so we identified this feeding behaviour as being problematic. To try to discourage feeding in this way we positioned the ledges beneath the feeder in such a way that it would be considerably more difficult for the average kākā to comfortably reach up enough to be able to cover the nozzle. Nevertheless, the larger kākā were still able to feed and obscure the nozzle. We were partially constrained in our attempts to discourage this feeding behaviour because the ledges could not be positioned so far below the feeder that the smaller kākā were unable to use the feeder. 

As a result, these issues for our object localisation process are expected, to a degree. We consider them acceptable given that the large majority of the dataset for which object localisation vastly improves the feature set for feature matching. 


\section{Feature Extraction}\label{sec:feature-extraction}

Feature extraction is composed of feature detection and feature description. These steps build upon the work in Section \ref{sec:object-localisation} to extract features from the region specified by kākā localisation.

\subsection{Feature Detection}

\begin{figure}[H]
  \centering
  \begin{minipage}[t]{0.48\textwidth}
    \includegraphics[width=\textwidth]{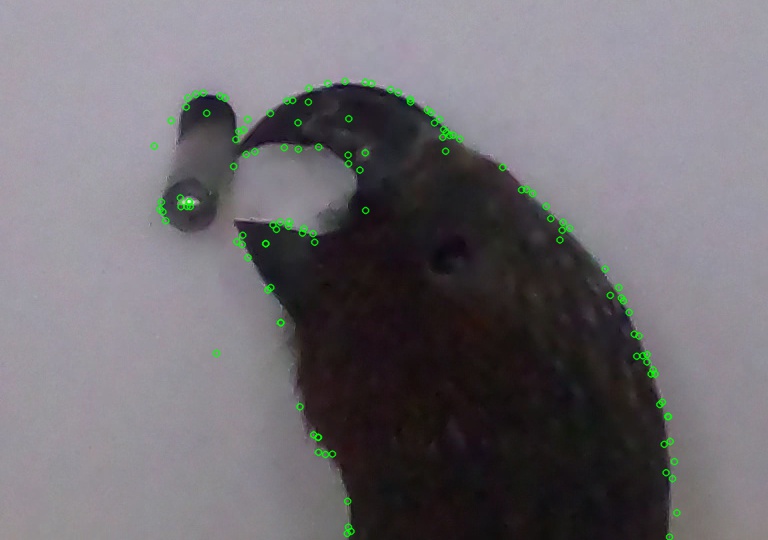}
    \caption{SIFT features detected without object localisation.}
     \label{fig:sift-features}
  \end{minipage}\hfill
  \begin{minipage}[t]{0.48\textwidth}
    \includegraphics[width=\textwidth]{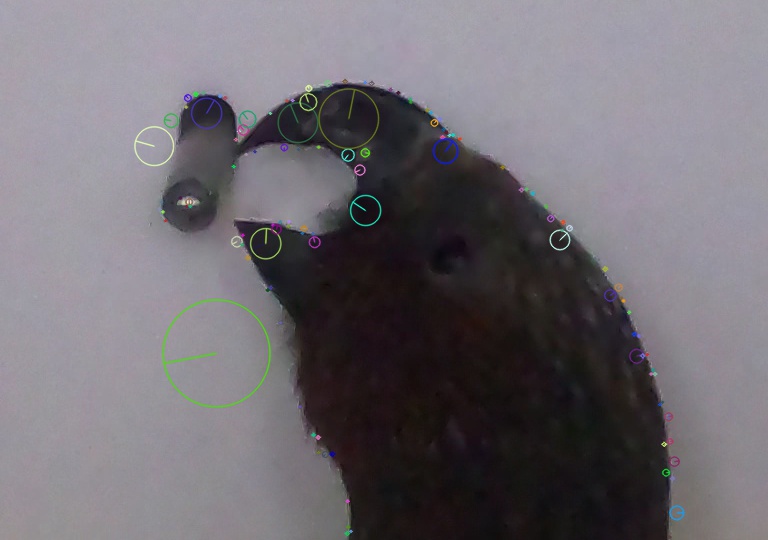}
    \caption{SIFT features detected without object localisation with size corresponding to detection scale.}
    \label{fig:sift-features-with-size}
  \end{minipage}
\end{figure}

With object localisation, we can create a localisation mask to filter features from the feature set. Feature detection is therefore as straight forward as applying SIFT feature detection and filtering keypoints according to the kākā localisation mask. We can see an example of feature detection without object localisation in Figure \ref{fig:sift-features} and Figure \ref{fig:sift-features-with-size}. Figure \ref{fig:sift-features} shows the keypoint features detected with SIFT and Figure \ref{fig:sift-features-with-size} shows the keypoint features with size corresponding to the scale at which the feature was detected. The majority of the features are detected on the edge of the bird, at points of interest on the beak and on the nozzle. Figure \ref{fig:sift-features-with-size} illustrates that features along the edges of the kākā and nozzle were detected in a small scale space while features within the nozzle, kākā, and in the background, are detected at larger scale spaces. 

\begin{figure}[H]
  \centering
  \begin{minipage}[t]{0.48\textwidth}
    \includegraphics[width=\textwidth]{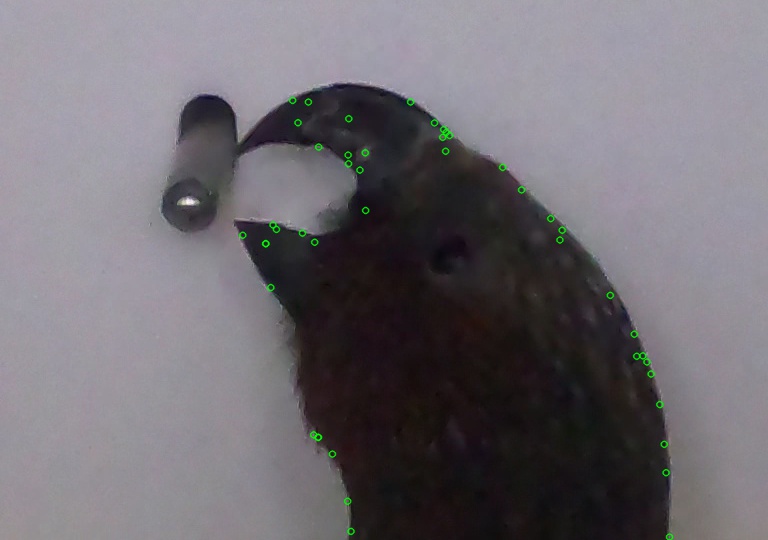}
    \caption{SIFT features detected and filtered by object localisation.}
     \label{fig:filtered-sift-features}
  \end{minipage}\hfill
  \begin{minipage}[t]{0.48\textwidth}
    \includegraphics[width=\textwidth]{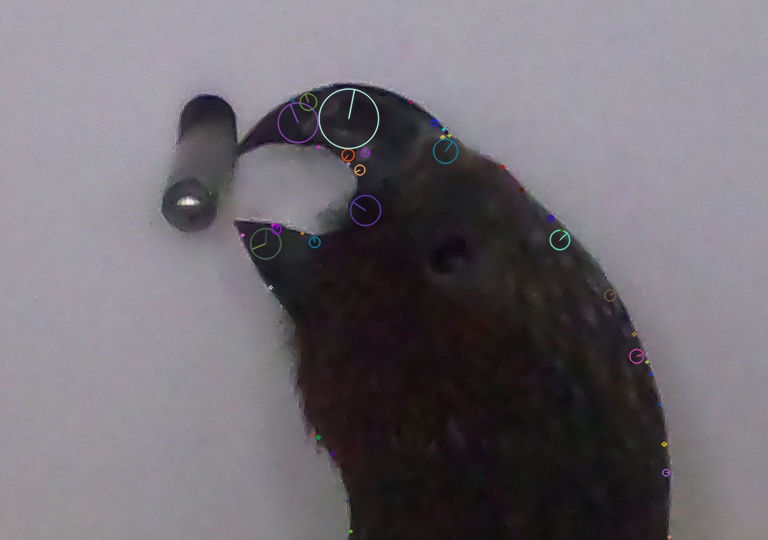}
    \caption{SIFT features detected and filtered by object localisation with size corresponding to detection scale.}
    \label{fig:filtered-sift-features-with-size}
  \end{minipage}
\end{figure}

When these features are filtered with the kākā localisation mask produced by work in Section \ref{sec:object-localisation}, the remaining features are shown in Figure \ref{fig:filtered-sift-features} and Figure \ref{fig:filtered-sift-features-with-size}. Figure \ref{fig:filtered-sift-features} shows the keypoint features detected with SIFT and filtered by object localisation and Figure \ref{fig:filtered-sift-features-with-size} shows the keypoint features with size corresponding to the scale at which the feature was detected, also filtered by a object localisation mask. These images show significantly fewer features and all the keypoint features are located on the kākā. No features on the nozzle pass through the localisation mask and there are fewer features detected along the edge of the kākā. The features removed along the edge of the kākā are low quality and are not always reliably discriminatory between individual kākā. As shown in Figure \ref{fig:filtered-sift-features-with-size}, most of the features detected at very small or very large scale spaces have also been removed, with the exception of one feature in the centre of the beak and several features along the back of the kākā's head. 

\subsection{Feature Description}

Descriptors for each feature are calculated in the same stage as feature detection so no additional computation or calculations are required to find the descriptors for each feature. Feature descriptors are calculated using the SIFT feature description method, outlined in Section \ref{sec:sift}.

\section{Feature Matching}


Feature matching takes the features extracted in Section \ref{sec:feature-extraction} and finds features that match between different images. This is the process which ultimately determines which images are matches matched with one another. 

There are a variety of feature matching methods available in a feature-based image matching approach. This section will apply the indirect matching methods outlined in Section \ref{sec:proposed-feature-matching} and qualitatively compare their results. Those matching methods are NN, NNDR and MNN. Each matching method produces a preliminary match set which is further improved by RANSAC mismatch removal. 

\begin{figure}[H]
\centering
  \includegraphics[width=\textwidth]{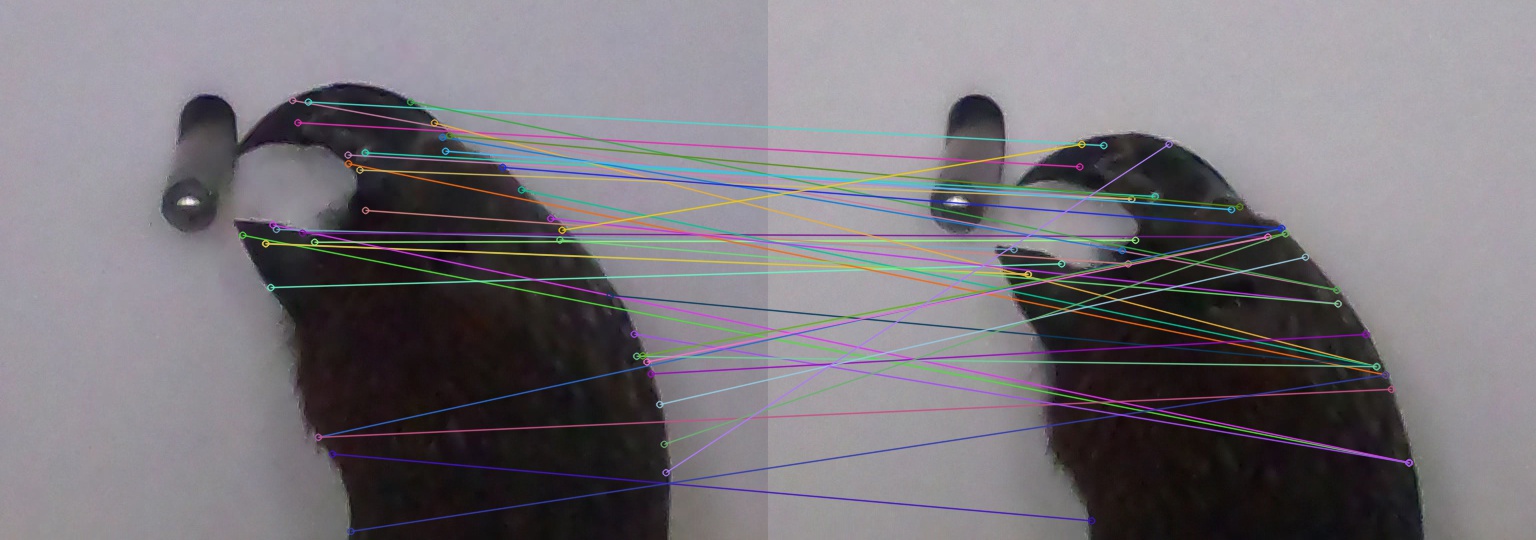}
  \caption{NN preliminary match set with example image pairing.}
  \label{fig:nn-without-ransac}
\end{figure}

\begin{figure}[H]
\centering
  \includegraphics[width=\textwidth]{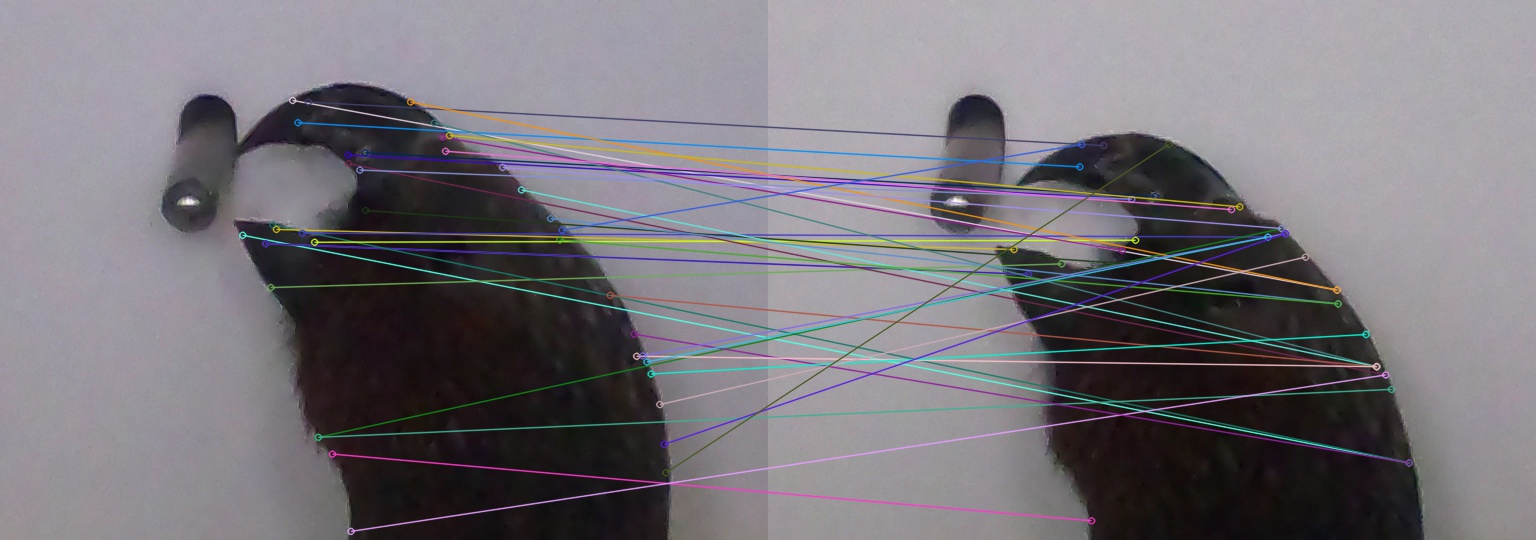}
  \caption{NNDR preliminary match set with example image pairing.}
  \label{fig:nndr-without-ransac}
\end{figure}

\begin{figure}[H]
\centering
  \includegraphics[width=\textwidth]{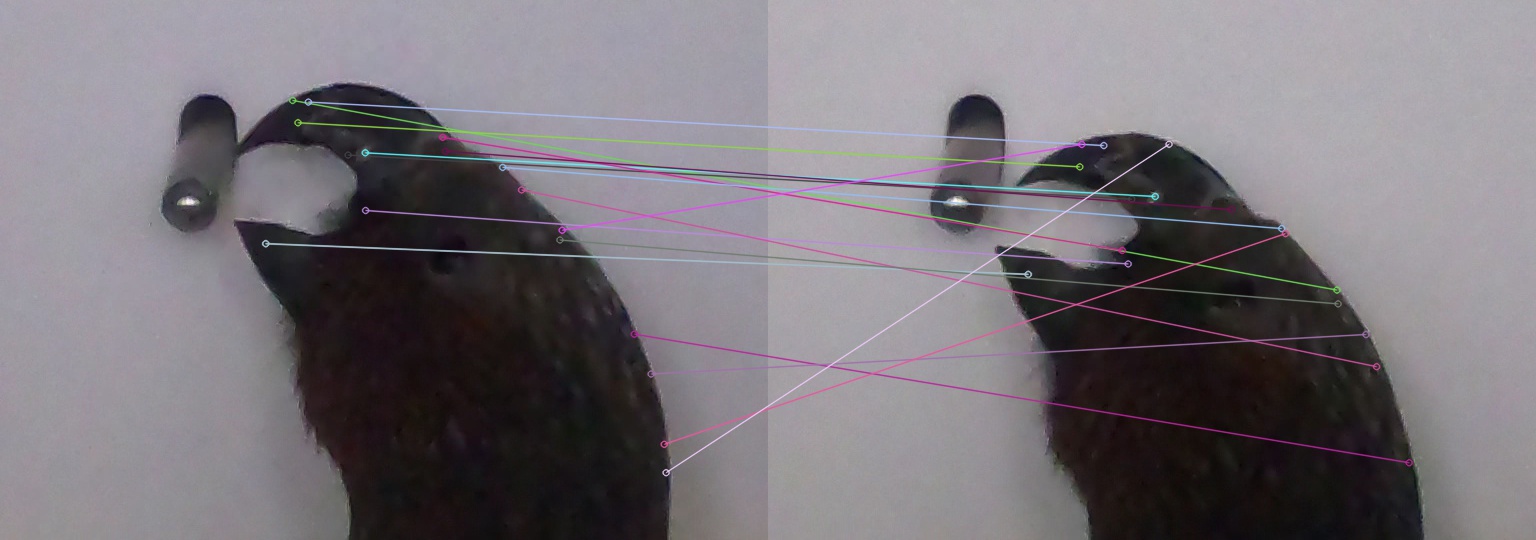}
  \caption{MNN preliminary match set with example image pairing.}
  \label{fig:mnn-without-ransac}
\end{figure}

In Figures \ref{fig:nn-without-ransac}, \ref{fig:nndr-without-ransac} and \ref{fig:mnn-without-ransac} we compare the preliminary match set produced by NN, NNDR and MNN without applying RANSAC mismatch removal with a true image pairing (where the two images contain the same kākā). Both nearest neighbour and nearest neighbour distance ratio identify 39 potential feature match candidates whereas mutual nearest neighbour identifies only 18 potential feature matches. This reflects the fact that MNN is a more selective matching strategy than the other two methods. It is perhaps surprising that NNDR identifies the same number of matches as NN because NNDR is more rigorous in selecting matches. Indeed, the matches identified by NN and NNDR are exactly the same matches. In a true image pairing such as this we would expect a high number of good feature matches, so the matches identified by NN must be of high enough quality that they also pass NNDR. 

\begin{figure}[H]
\centering
  \includegraphics[width=\textwidth]{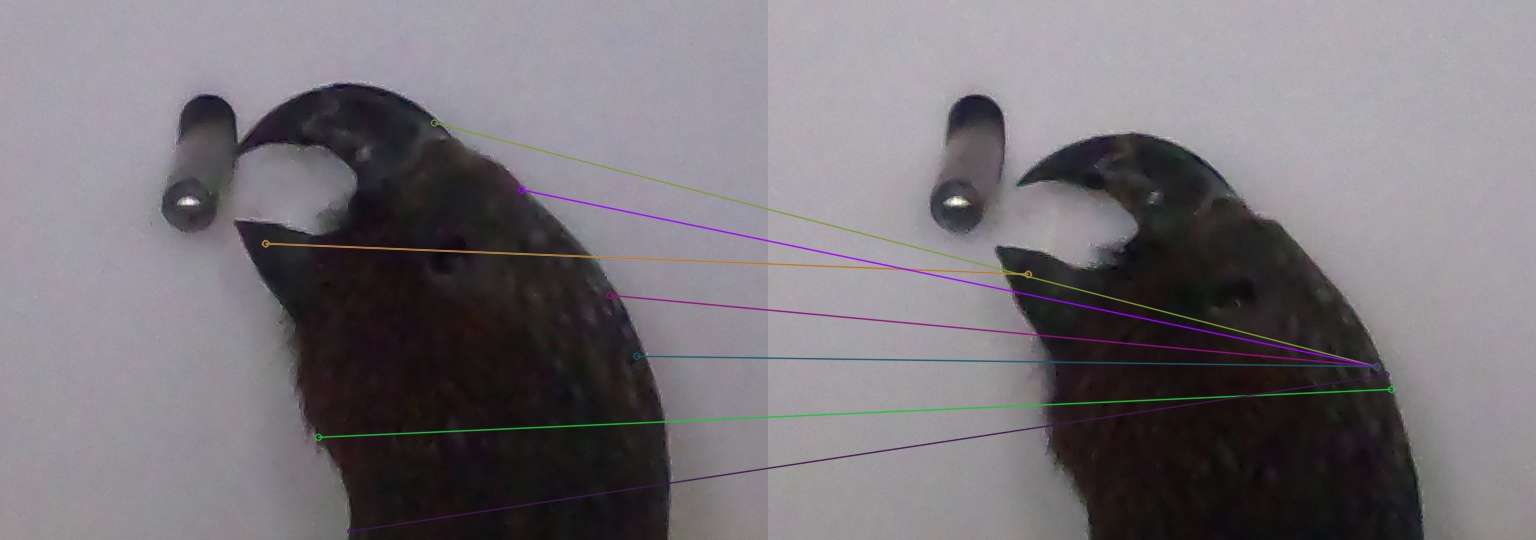}
  \caption{NN preliminary match set with example image pairing.}
  \label{fig:nn-with-ransac}
\end{figure}

\begin{figure}[H]
\centering
  \includegraphics[width=\textwidth]{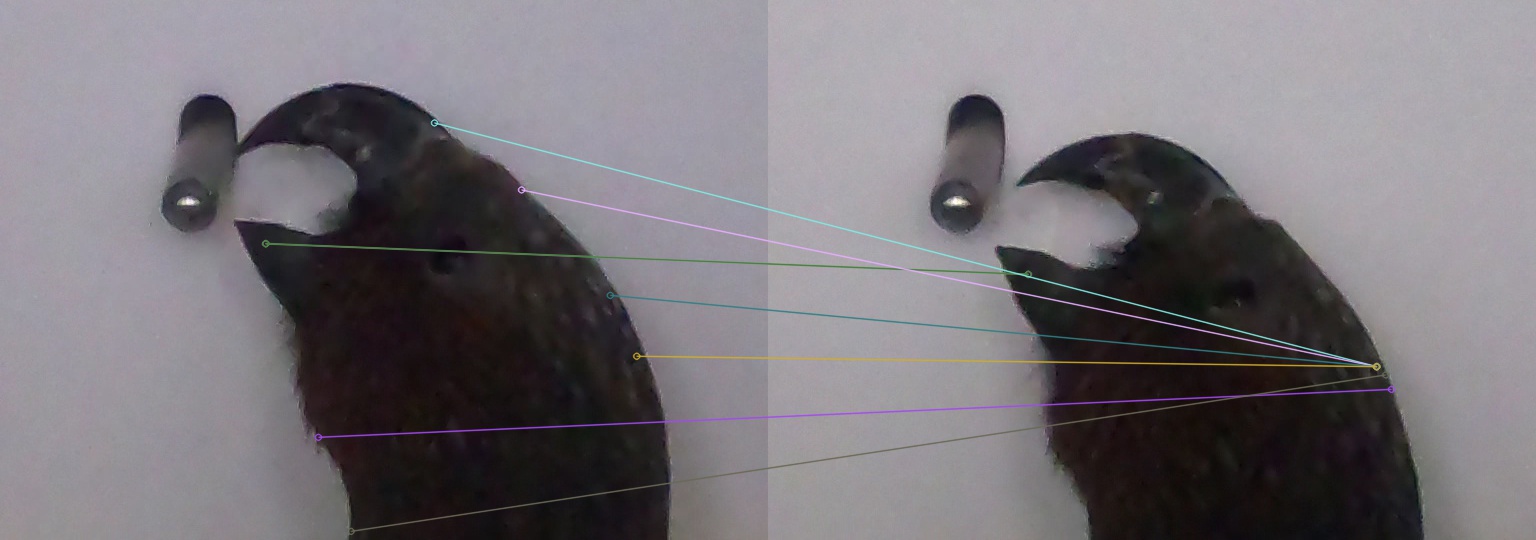}
  \caption{NNDR preliminary match set with example image pairing.}
  \label{fig:nndr-with-ransac}
\end{figure}

\begin{figure}[H]
\centering
  \includegraphics[width=\textwidth]{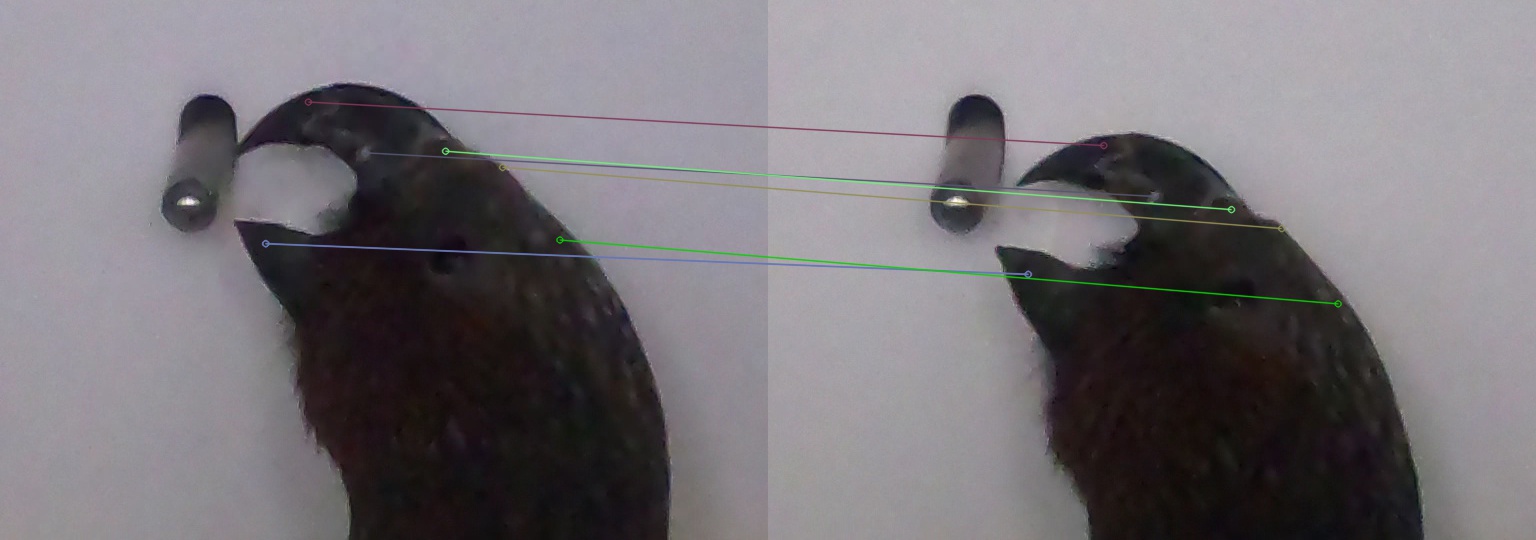}
  \caption{MNN preliminary match set with example image pairing.}
  \label{fig:mnn-with-ransac}
\end{figure}

In Figures \ref{fig:nn-with-ransac}, \ref{fig:nndr-with-ransac}, and \ref{fig:mnn-with-ransac} we can see the impact of the RANSAC mismatch removal on the feature match set. Once again, the feature match set with NN and NNDR are the exact same, containing seven feature matches between the two images. MNN and RANSAC combine to reduce the number of feature matches to six. Visually, the feature matches identified by MNN are much more convincing. Features matched are also located principally on the beak of the kākā where as the weaker matches produced by NN and NNDR using features along the outline of the kākā. 

The main reason for the disparity in performance of MNN in comparison to NN and NNDR is that MNN enforces one to one correspondence between feature matches. This prevents feature matches like those shown in Figures \ref{fig:nn-with-ransac} and \ref{fig:nndr-with-ransac}, where seven features in the image on the left are matched to only four features on the right. Not requiring one-to-one correspondence between features in matches can help find more potentially true matches, because it is not exclusive. However, if we would expect two images to match then features should match in a one-to-one nature.

It is also interesting to note that after RANSAC, NN and NNDR feature sets have only one more feature match than MNN. So, the larger preliminary feature match set of NN and NNDR are really just identifying more matches that will be removed after RANSAC.

From this stage, we conclude that MNN is the most effective feature matching technique for identifying individual kākā because it demands one-to-one correspondence between features to match. This is a property which is much more likely to be met by images of the same individual kākā because features between those images are a true match. That is, the features represent the same relevant and identifying point on the kākā. 

\section{Image Matching}\label{sec:image-matching}

From feature matches between images, we can calculate the best overall match for an image. The best indicator for the quality of an image match is the number of feature matches between the two images, but the average distance between the descriptors of those features matches is also a useful metric. 

\begin{figure}[H]
\caption{Best matches according the number of descriptor matches between images.} 
\centering 
\begin{tabular}{c c}
\includegraphics[width=.48\textwidth,valign=m]{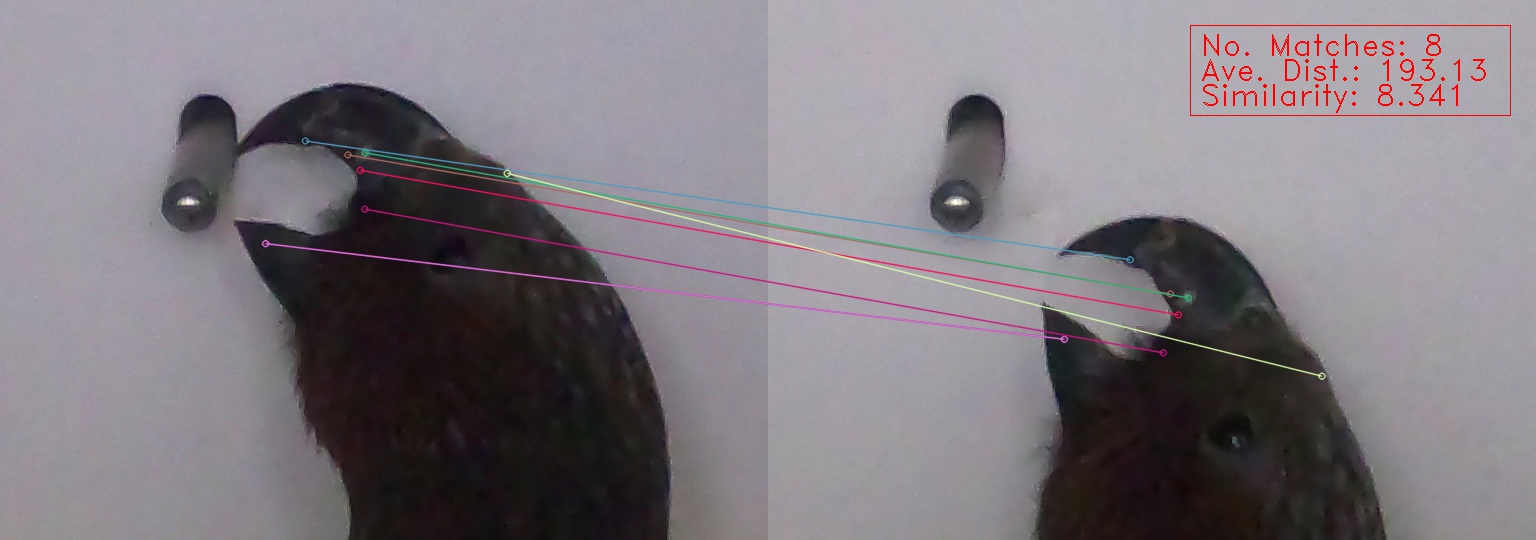} & \includegraphics[width=.48\textwidth,valign=m]{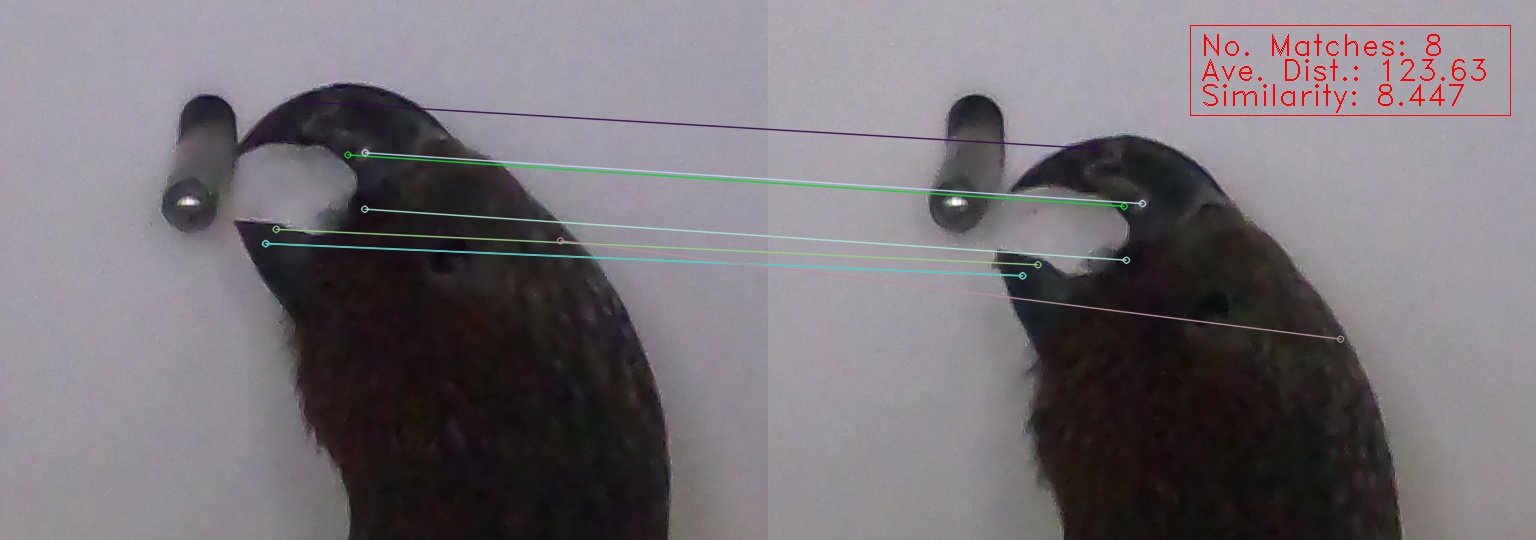} \\
\textbf{(a)} & \textbf{(b)}\\
\includegraphics[width=.48\textwidth,valign=m]{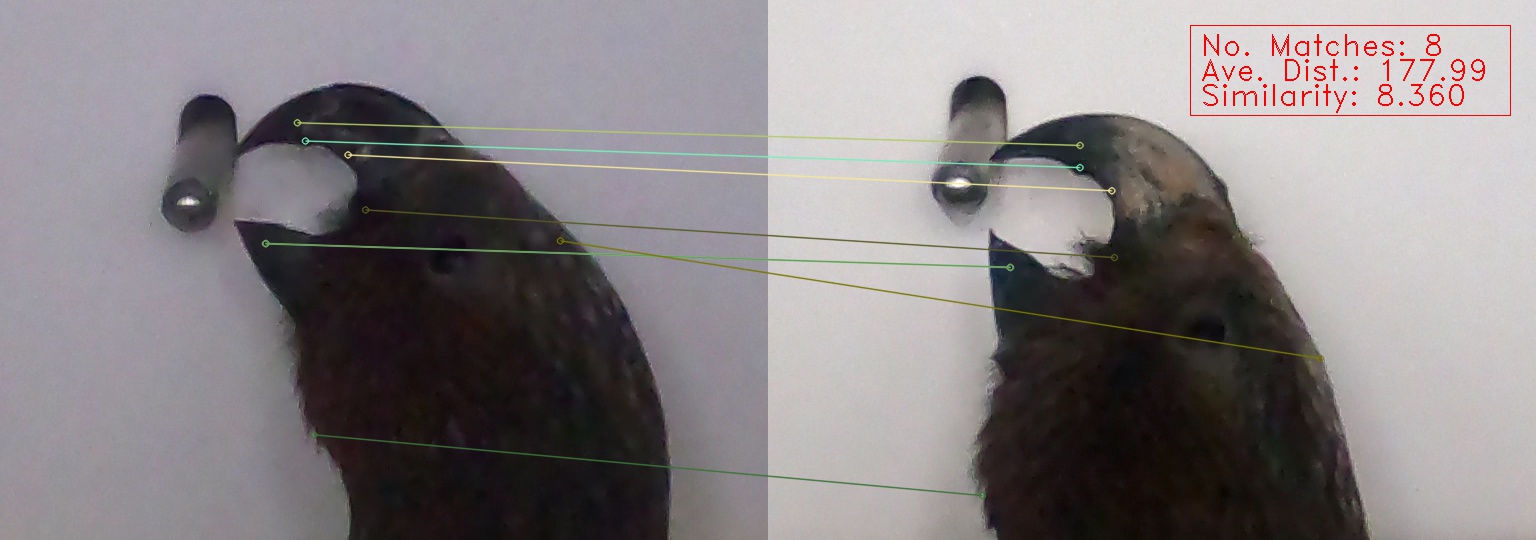} &
\includegraphics[width=.48\textwidth,valign=m]{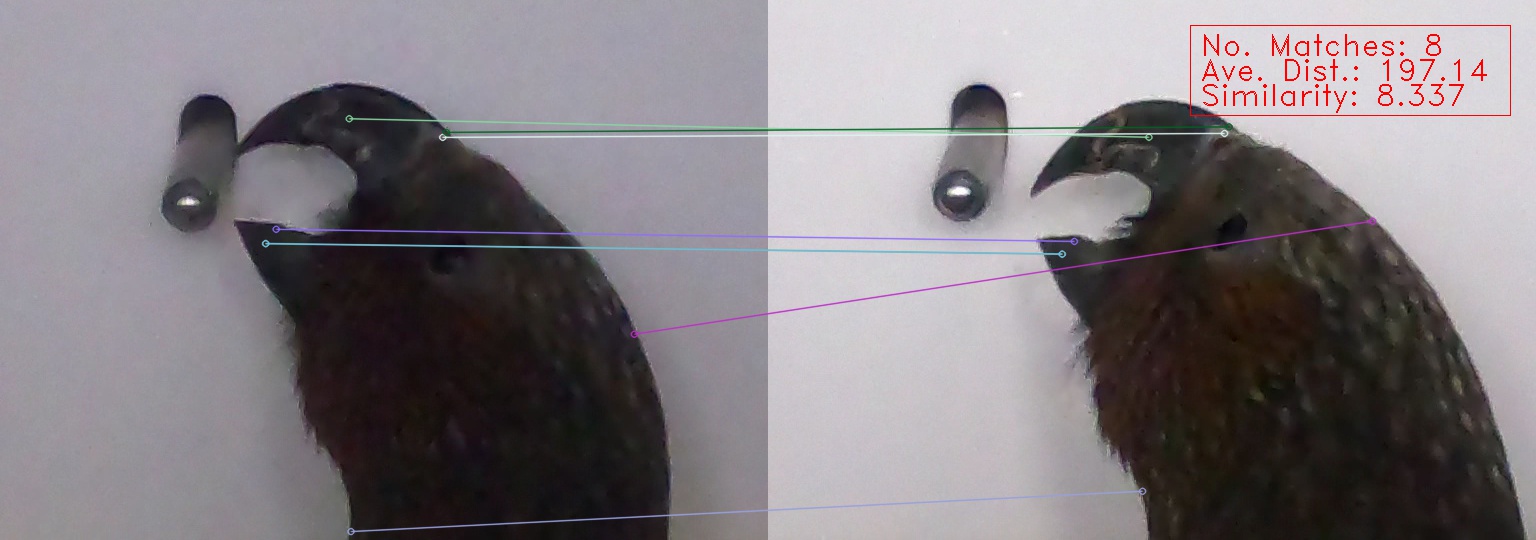} \\
\textbf{(c)} & \textbf{(d)}\\
\includegraphics[width=.48\textwidth,valign=m]{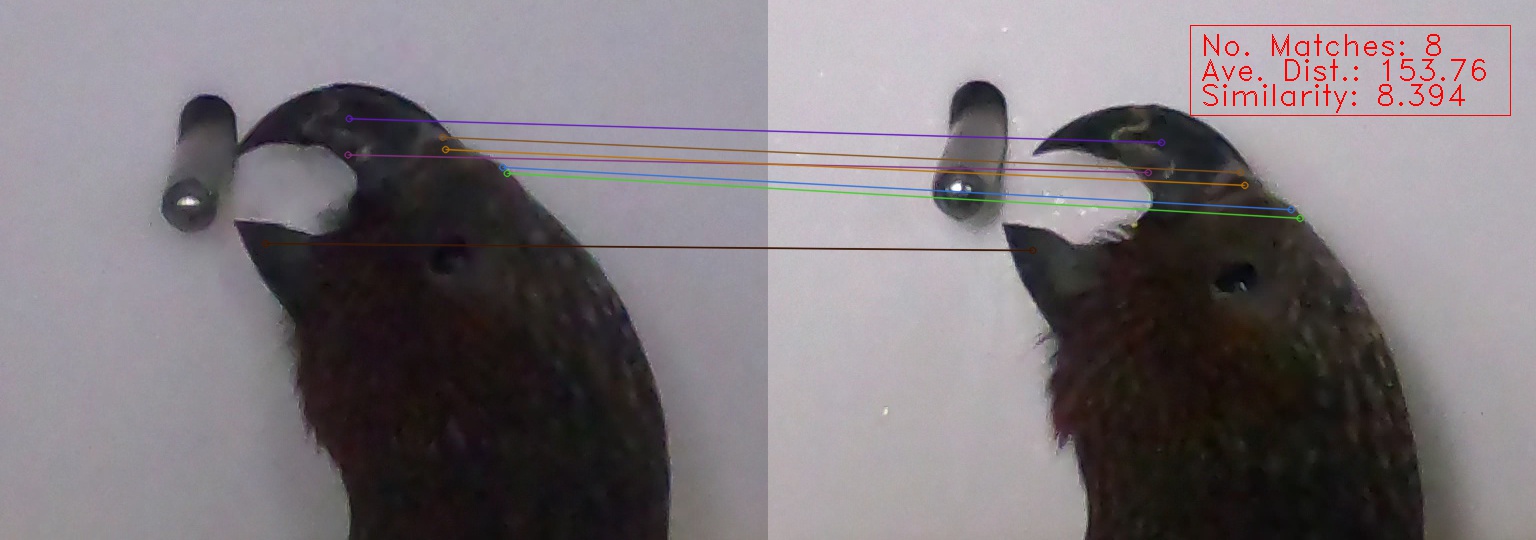} & \includegraphics[width=.48\textwidth,valign=m]{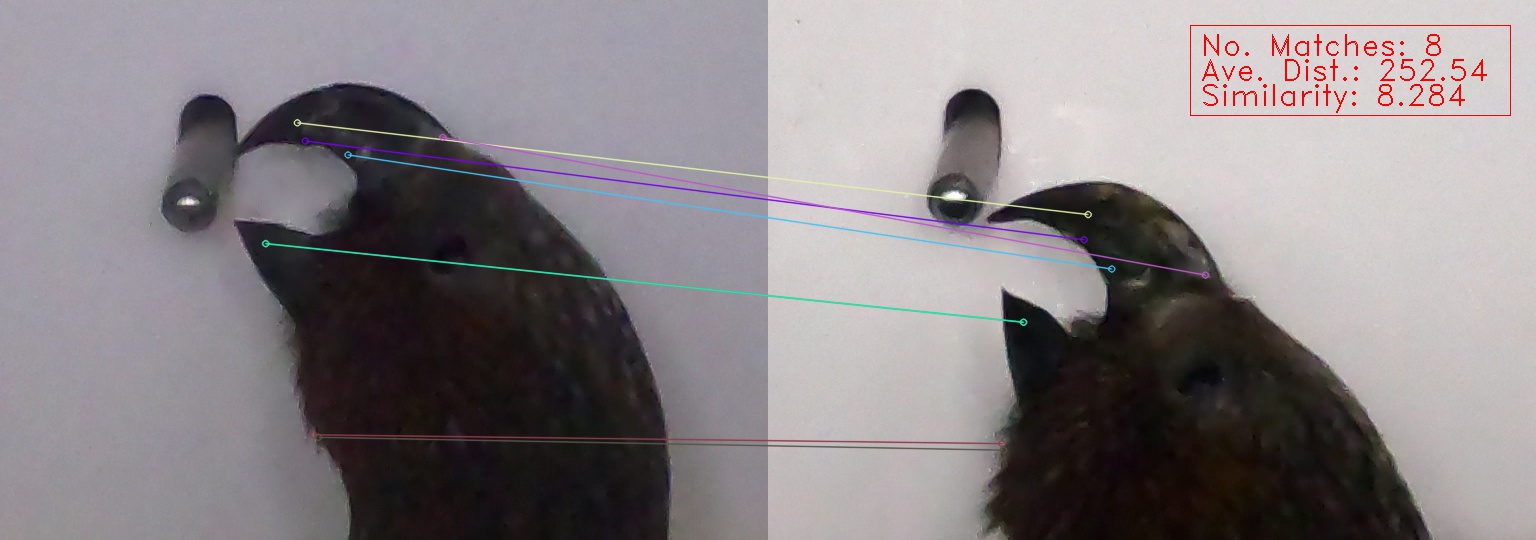}\\
\textbf{(e)} & \textbf{(f)}\\
\includegraphics[width=.48\textwidth,valign=m]{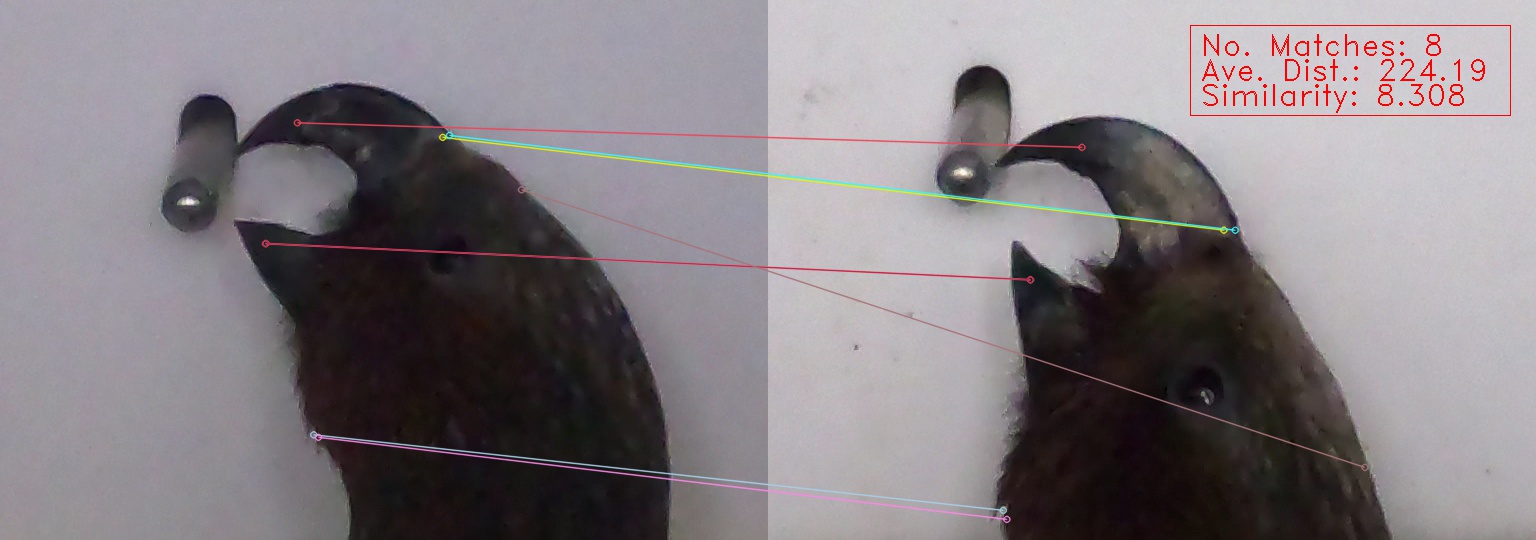} & \includegraphics[width=.48\textwidth,valign=m]{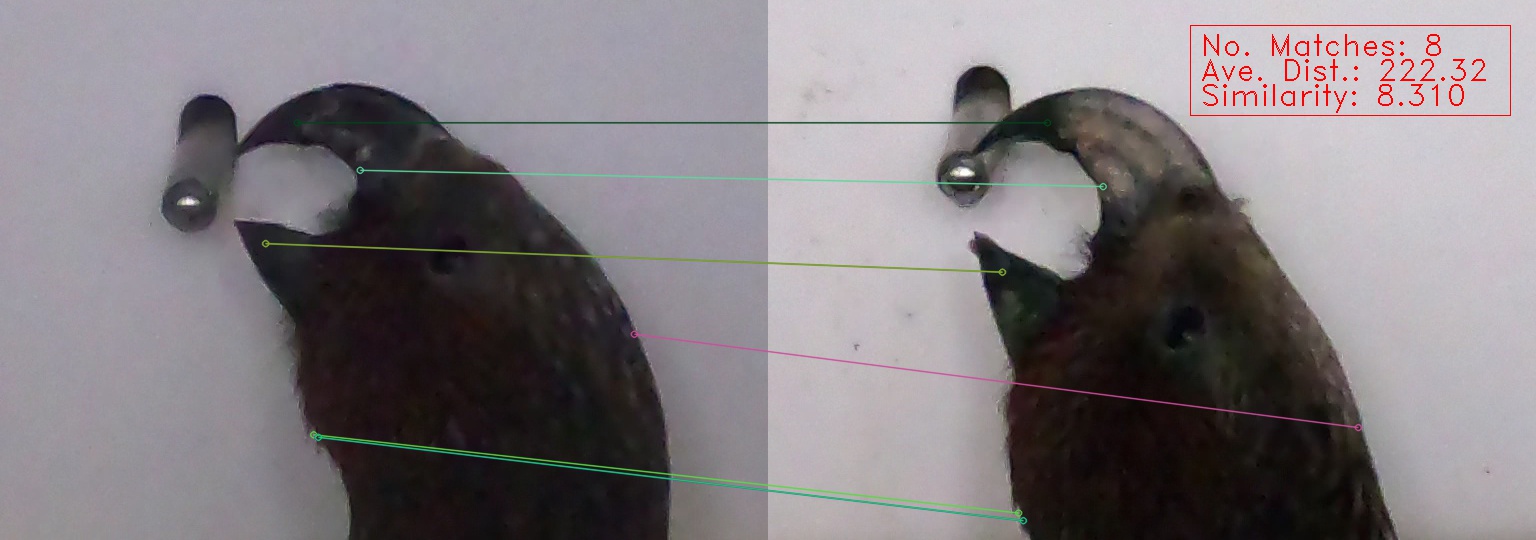} \\
\textbf{(g)} & \textbf{(h)}\\
\end{tabular}
\label{fig:best-matches-max-matches}
\end{figure}

\begin{figure}[H]
\centering
  \includegraphics[width=\textwidth]{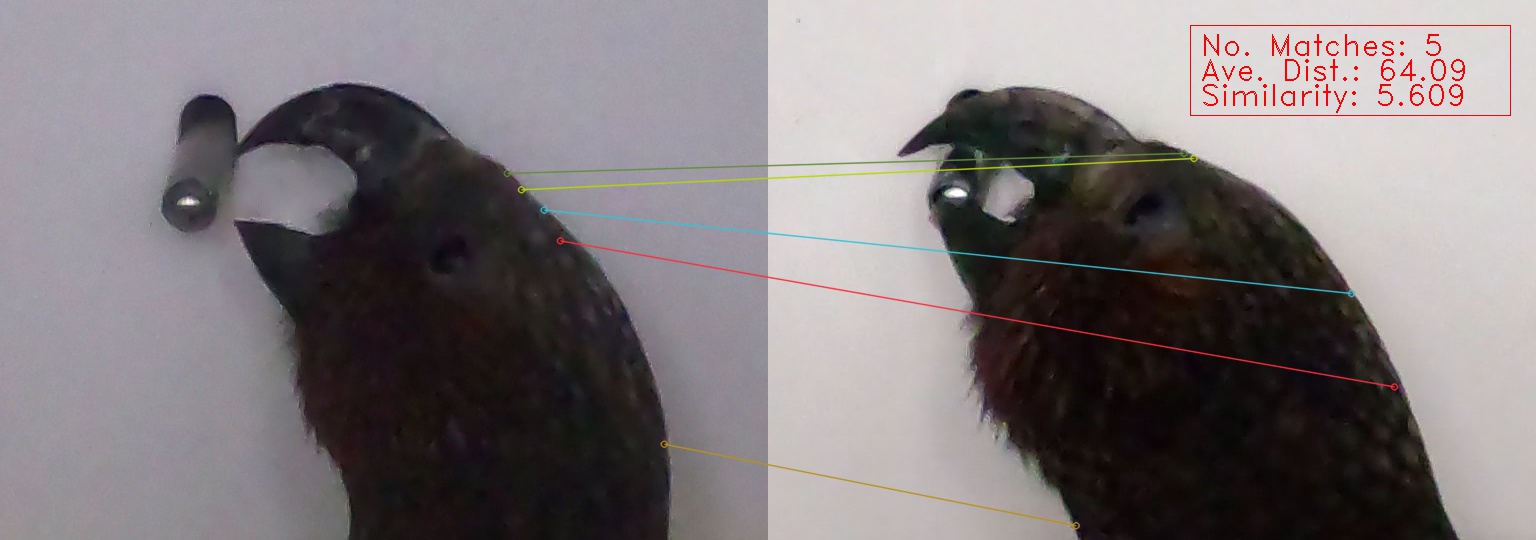}
  \caption{Best match according to average distance between descriptor matches.}
  \label{fig:best-match-average-distance}
\end{figure}

Figure \ref{fig:best-matches-max-matches} shows the best matches according to which images have the highest number of feature matches while Figure \ref{fig:best-match-average-distance} shows the best match according to the average distance between matching feature descriptors. Already, in the fact that there are eight different images identified as the best match according to the number of features, we can see the need for further distinction between matches. Additionally, Figure \ref{fig:best-match-average-distance} illustrates that average distance cannot be used in isolation. By more closely examining the matches in Figure \ref{fig:best-matches-max-matches}, we observe that all matches except for \textbf{(c)}, \textbf{(g)}, and \textbf{(h)} identify a correct match. \textbf{(g)} and \textbf{(h)} have the highest average distance between feature descriptors out of all matches and only \textbf{(a)} and \textbf{(e)} have a higher average distance than \textbf{(c)}. This is probably because the kākā in \textbf{(c)} is positioned more similarly than \textbf{(a)} or \textbf{(e)}.

\begin{figure}[H]
\centering
  \includegraphics[width=\textwidth]{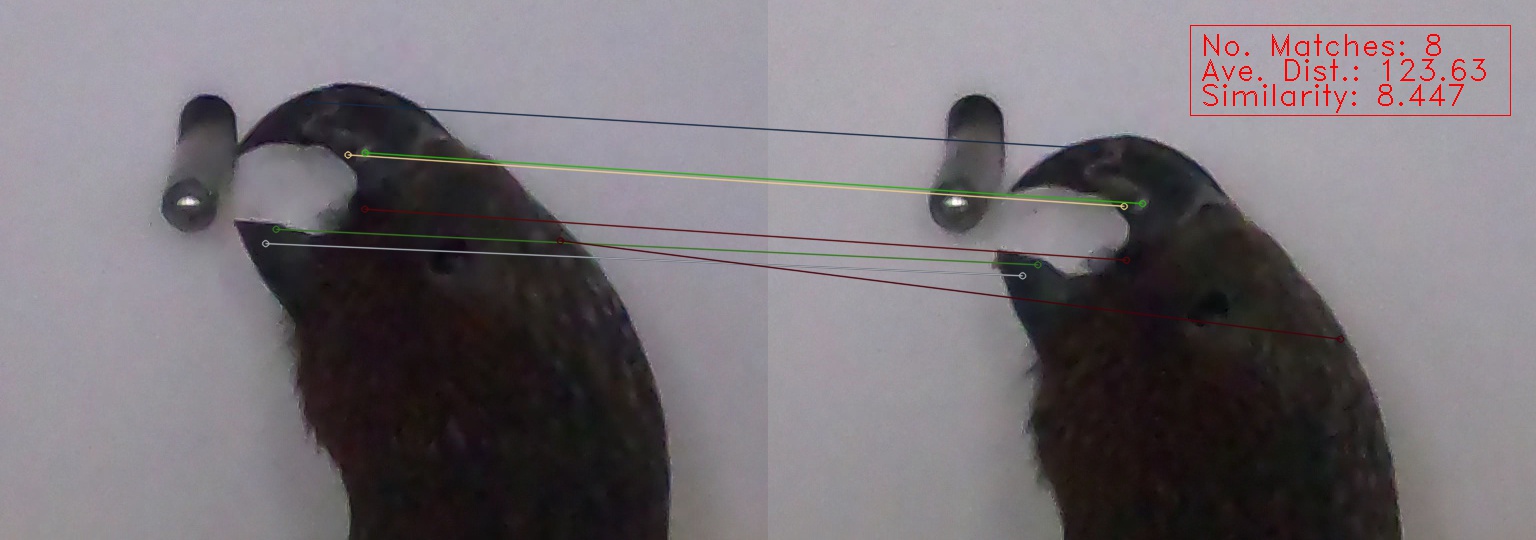}
  \caption{Best match according to the similarity measure \(S\).}
  \label{fig:best-match-similarity-score}
\end{figure}

Although it is not a perfect metric, using average distance as an additional discriminator between matches allows us to order images with the same number of feature matches. This relative weighting of matches and average distance informs our similarity measure \(S\), as outlined in Section \ref{sec:similarity-measure} and calculated by the set of distances between descriptor matches \(D\):

\[ S(D) = |D| + \frac{1}{1 + \sum_{n=1}^{D} \frac{d_n}{|D|}}\]
The overall best image match out of those in Figure \ref{fig:best-matches-max-matches}, according to the similarity measure \(S\), is \textbf{(a)}, shown again in Figure \ref{fig:best-match-similarity-score}. The similarity measure is denoted in the top right of Figure \ref{fig:best-matches-max-matches}.

\chapter{Results}\label{C:results}

The final image matching function uses a localisation mask with SIFT feature extraction to detect features within each image, before using mutual nearest neighbour matching and RANSAC to match features with potential image matches. The similarity score outlined in Section \ref{sec:similarity-measure} is used to rank images by similarity in order to find each images best match. 

The results gathered from evaluating the top \(X\) best matches of the image matching function with the labelled set, for \(X\) values of 1, 2, and 3, are shown in Table 5.1, Table 5.2 and Table 5.3 respectively. 

Each label, corresponding to a kākā band combination is displayed on the left. An image is considered to be classified correctly if at least one of the top \(X\) best matches for that image shares the same label. A match is categorised as incorrect if none of the top \(X\) best matches for that image share the same label. The total column refers to the total number of images in that label set and the accuracy indicates the percentage of images in the label set where at least one of the top \(X\) matches is correct. The overall totals for each column are displayed in the bottom row. 



\begin{table}[H]
\caption{Best Match Image Function Evaluation Results} 
\centering 
\begin{tabular}{c c c c c} 
\hline 
\textbf{Label} & \textbf{Correct} & \textbf{Incorrect} & \textbf{Total} & \textbf{Accuracy} \\ 
\hline
Lime-PurpleBlue & 96 & 43 & 139 & 0.6906\\ [1ex]
LimePurple-Green & 65 & 37 & 102 & 0.6373 \\[1ex]
Orange-RedSilver & 191 & 11 & 202 & 0.9455\\[1ex]
PurpleRed-Red & 49 & 8 & 57 & 0.8596\\ [1ex]
WhiteSilver-Pink & 86 & 53 & 139 & 0.6187\\ [1ex]
Yellow-GreenPurple & 8 & 1 & 9 & 0.8889\\ [1ex]
YellowPurple-Yellow & 128 & 19 & 147 & 0.8707\\ [1ex]
\hline 
 &  623 & 172 & 795 & 0.7836\\
\hline 
\end{tabular}
\label{table:results-1}
\end{table}


\begin{table}[H]
\caption{Top Two Matches Image Function Evaluation Results} 
\centering 
\begin{tabular}{c c c c c} 
\hline 
\textbf{Label} & \textbf{Correct} & \textbf{Incorrect} & \textbf{Total} & \textbf{Accuracy} \\ 
\hline
Lime-PurpleBlue & 113 & 36 & 139 & 0.8130\\ [1ex]
LimePurple-Green & 79 & 23 & 102 & 0.7745 \\[1ex]
Orange-RedSilver & 198 & 4 & 202 & 0.9802\\[1ex]
PurpleRed-Red & 52 & 5 & 57 & 0.9123 \\ [1ex]
WhiteSilver-Pink & 105 & 34 & 139 & 0.7554\\ [1ex]
Yellow-GreenPurple & 9 & 0 & 9 & 1.000000\\ [1ex]
YellowPurple-Yellow & 136 & 11 & 147 & 0.9252\\ [1ex]
\hline 
 &  692 & 102 & 795 & 0.8704\\
\hline 
\end{tabular}
\label{table:results-2}
\end{table}


\begin{table}[H]
\caption{Top Three Matches Image Function Evaluation Results} 
\centering 
\begin{tabular}{c c c c c} 
\hline 
\textbf{Label} & \textbf{Correct} & \textbf{Incorrect} & \textbf{Total} & \textbf{Accuracy} \\ 
\hline
Lime-PurpleBlue & 118 & 31 & 139 & 0.8490 \\ [1ex]
LimePurple-Green & 81 & 21 & 102 & 0.7941 \\[1ex]
Nothing-Blue & 148 & 41 & 189 & 0.7831 \\[1ex]
Orange-RedSilver & 200 & 2 & 202 & 0.9901\\[1ex]
PurpleRed-Red & 54 & 3 & 57 & 0.9474 \\ [1ex]
WhiteSilver-Pink & 110 & 29 & 139 & 0.7914 \\ [1ex]
Yellow-GreenPurple & 9 & 0 & 9 & 1.000000\\ [1ex]
YellowPurple-Yellow & 138 & 9 & 147 & 0.9388\\ [1ex]
\hline 
 &  710 & 85 & 795 & 0.8931\\
\hline 
\end{tabular}
\label{table:results-3}
\end{table}

\section{Quantitative Analysis}

In Table \ref{table:results-1}, we can see that the best match for each image in the labelled subset was an accurate match in 78.36\% of cases. In Table \ref{table:results-2} and \ref{table:results-3}, results indicate that this accuracy rate increased to 87.04\% and then 89.31\% when we consider the best two and three matches, respectively. 

The evaluation accuracy in Table \ref{table:results-3} is a good overall indicator of the effectiveness of the image matching method, because the differences between the similarity scores for the top matches can be so close. As shown in Section \ref{sec:image-matching}, average distance between feature match descriptors is not a perfect method for choosing between matches with the same number of feature matches.If an accurate match was not found in the top three matches for a given image then we would be doubtful that the similarity score could be reliably used to inform a similarity network aiming to cluster individual kākā. With an accuracy of 89.31\% when considering the top three matches, we can assume that image matching would be very useful for informing a similarity network to identify individual kākā. It is even more encouraging that accuracy increases so much between \(X = 1\) and \(X = 2\), because it means that 40.11\% of images that were not accurately matched by their best match, were correctly matched by the next closest match. 

Of particular note in the accuracy scores among the labelled images, the kākā with the Orange-RedSilver band combination achieved an accuracy of 94.55\% accuracy for its best match. When the top two matches are considered, that accuracy increases to 98.02\%, and when the top three matches are considered, the accuracy increases further still to 95.54\%. Such high accuracy is potentially due to the higher total number of images with that label, because it means there are more possible correct matches. 

Nevertheless, the image matching function achieves an accuracy greater than 90\% when we consider three matches, for four separate labels have varying numbers of total images. Those labels are Yellow-GreenPurple, YellowPurple-Yellow, PurpleRed-Purple, and Orange-RedSilver in ascending order of most images with that label. 

The Yellow-GreenPurple label has only nine images yet accurate matches are found for all those images when we consider even just the best \textit{two} matches. Indeed, it might seem strange to include a label in the evaluation set when it has so few images, but this gives a good indicator of the sensitivity of the image matching function. That all images in the Yellow-GreenPurple label found an accurate match suggests that the image matching process is very sensitive to small differences between the kākā. 

The lowest accuracy score among the eight labels is for the WhiteSilver-Pink label, with an accuracy of 61.87\% for images with that label. However, among all labels, the total number of correct matches increases by the most as the number of matches considered increases from one to three. When the top two matches are considered, an additional 19 images are classified correctly and the accuracy increases to 75.54\%. When the top three matches are considered another 5 images find an accurate match and the overall accuracy on the label increases to 79.14\%. Lime-PurpleBlue and  LimePurple-Green labels see quantitatively similar increases in accuracy of approximately 16\%. 


\section{Qualitative Analysis}

\begin{figure}[H]
\caption{Images with the YellowPurple-Yellow label without an accurate match.} 
\centering 
\begin{tabular}{c c c}
\includegraphics[width=.3\textwidth,valign=m]{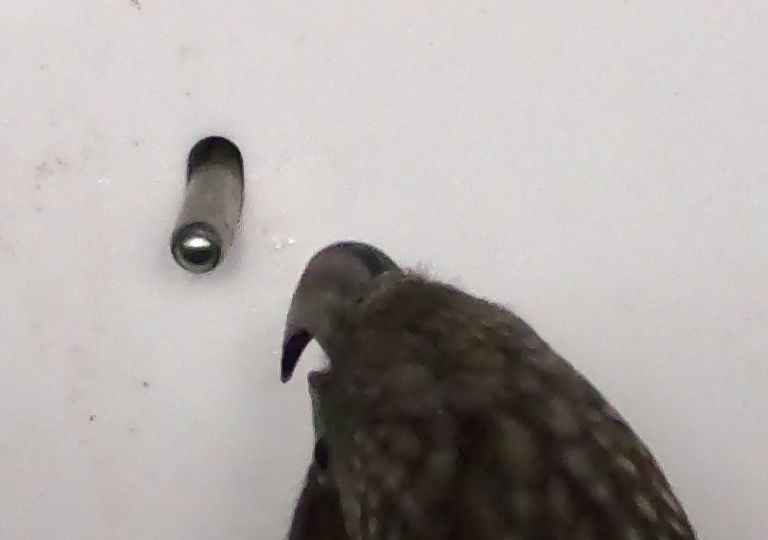} & \includegraphics[width=.3\textwidth,valign=m]{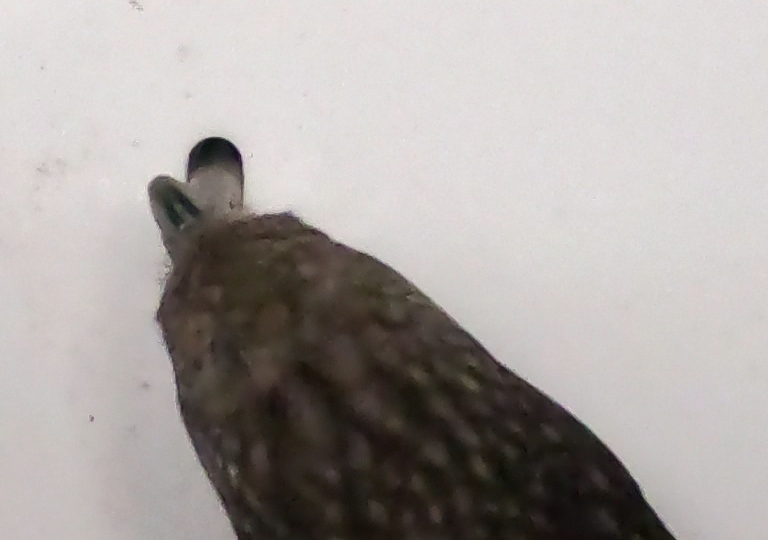} & \includegraphics[width=.3\textwidth,valign=m]{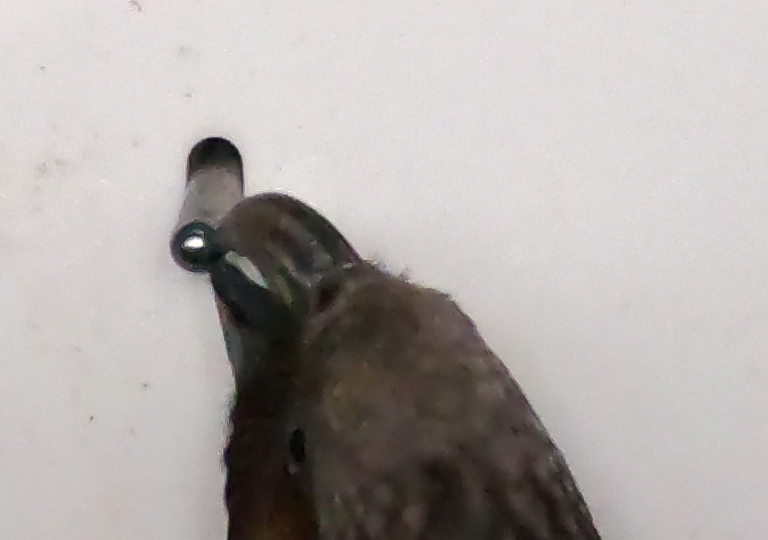}\\
\textbf{(a)} & \textbf{(b)} & \textbf{(c)} \\
\includegraphics[width=.3\textwidth,valign=m]{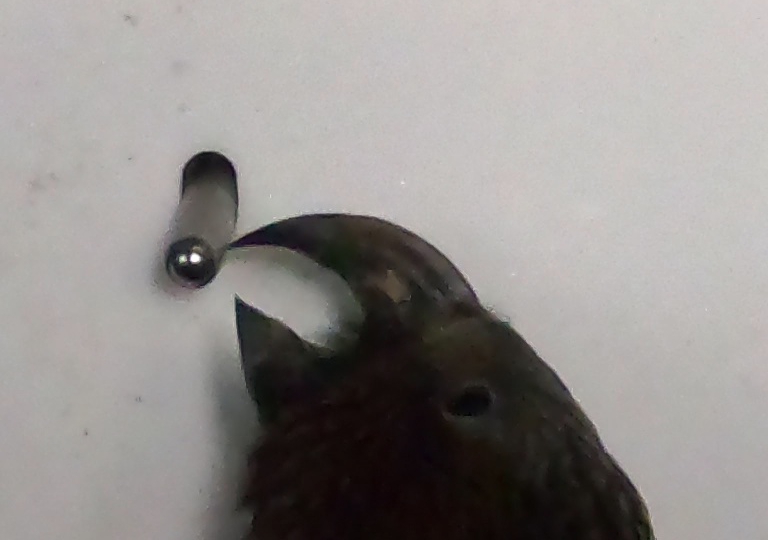} & \includegraphics[width=.3\textwidth,valign=m]{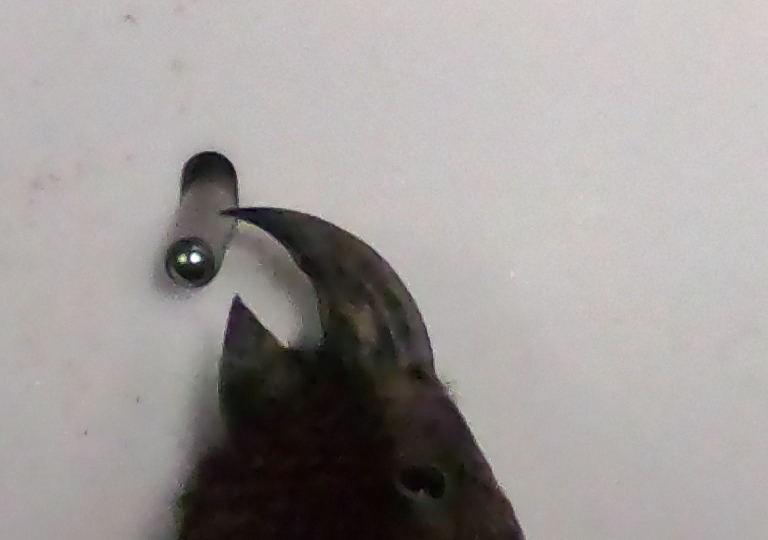} & \includegraphics[width=.3\textwidth,valign=m]{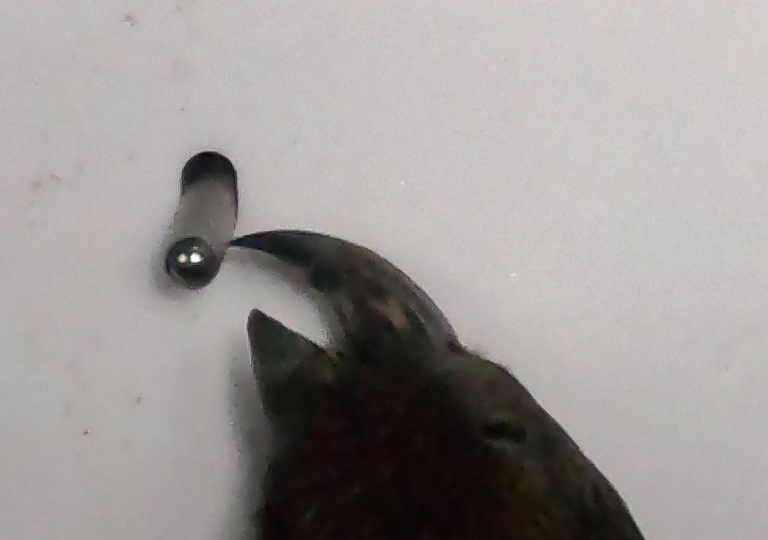}\\
\textbf{(d)} & \textbf{(e)} & \textbf{(f)} \\
\includegraphics[width=.3\textwidth,valign=m]{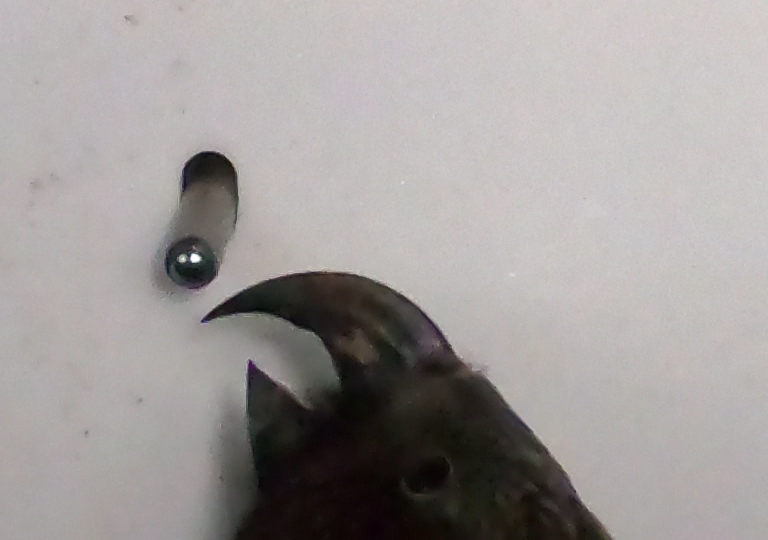} & \includegraphics[width=.3\textwidth,valign=m]{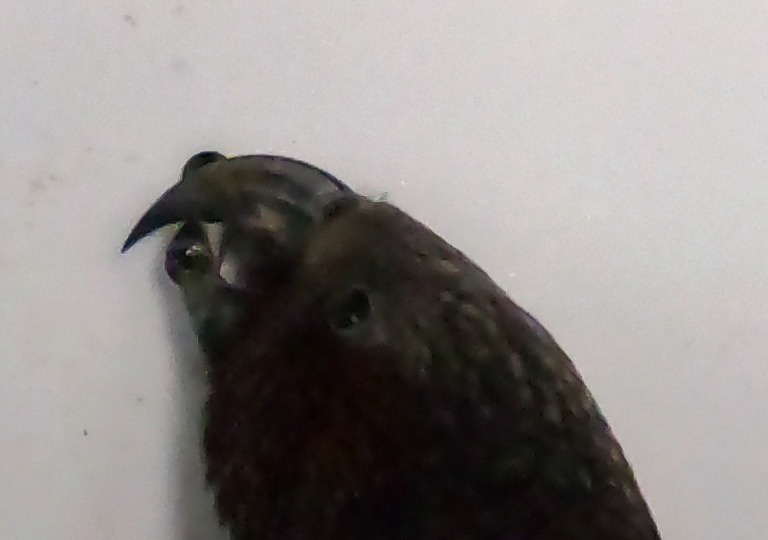} & \includegraphics[width=.3\textwidth,valign=m]{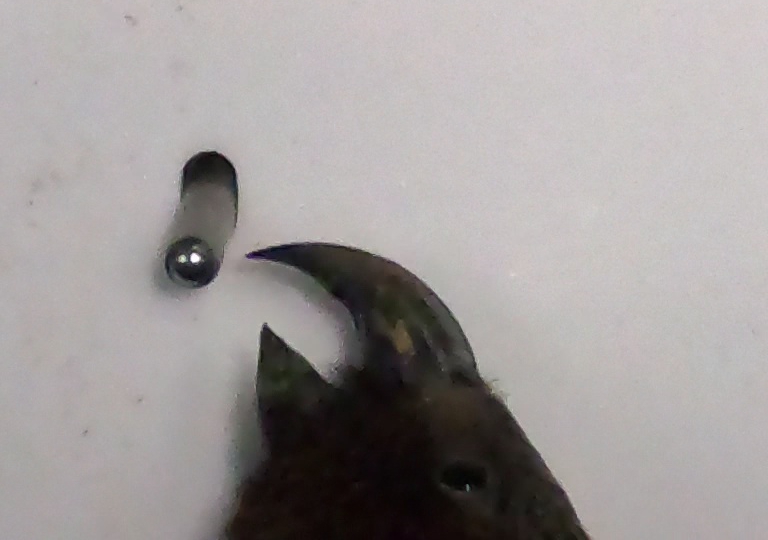}\\
\textbf{(g)} & \textbf{(h)} & \textbf{(i)} \\
\end{tabular}
\label{fig:no-accurate-matches-9}
\end{figure}

To understand why the image matching process might fail to find an accurate match for an image, let us explore the nine image with the YellowPurple-Yellow label that image matching did not find an accurate match for. These images are shown in Figure \ref{fig:no-accurate-matches-9}. For images \textbf{(a)}, \textbf{(b)} and \textbf{(c)} we can see that the quality of the image is quite low. The beak is not fully visible in any of them and the kākā is not in profile. On the contrary, the back of the head is the most prominent part of the kākā. These images would likely have been close to being removed during preprocessing. Nevertheless, it's important to see how the image matching process handles bad data.

For \textbf{(h)}, the beak of the nozzle is obscuring the nozzle. This means that in the object detection phase of feature extraction, as discussed in Section 5, features in the part of the beak obscuring the nozzle will be filtered out. Therefore, the lack of accurate matches may be due to a lack of quality features with which to provide meaningful information about the image for matching. After all, the most distinguishing feature for each of the kākā is their beak. 

For images \textbf{(d)}, \textbf{(e)}, \textbf{(f)},  \textbf{(g)} and \textbf{(i)}, the images are not so obviously of poor quality. The positioning of the birds is acceptable because the beak is fully visible. However, the kākā are in the upward motion of feeding, leading to a blurred effect of the details in the beaks. Additionally, the beak is the only visible part of the kākā. Therefore, the issue may be that there are not enough important features extracted.

\begin{figure}[H]
\centering
  \includegraphics[width=\textwidth]{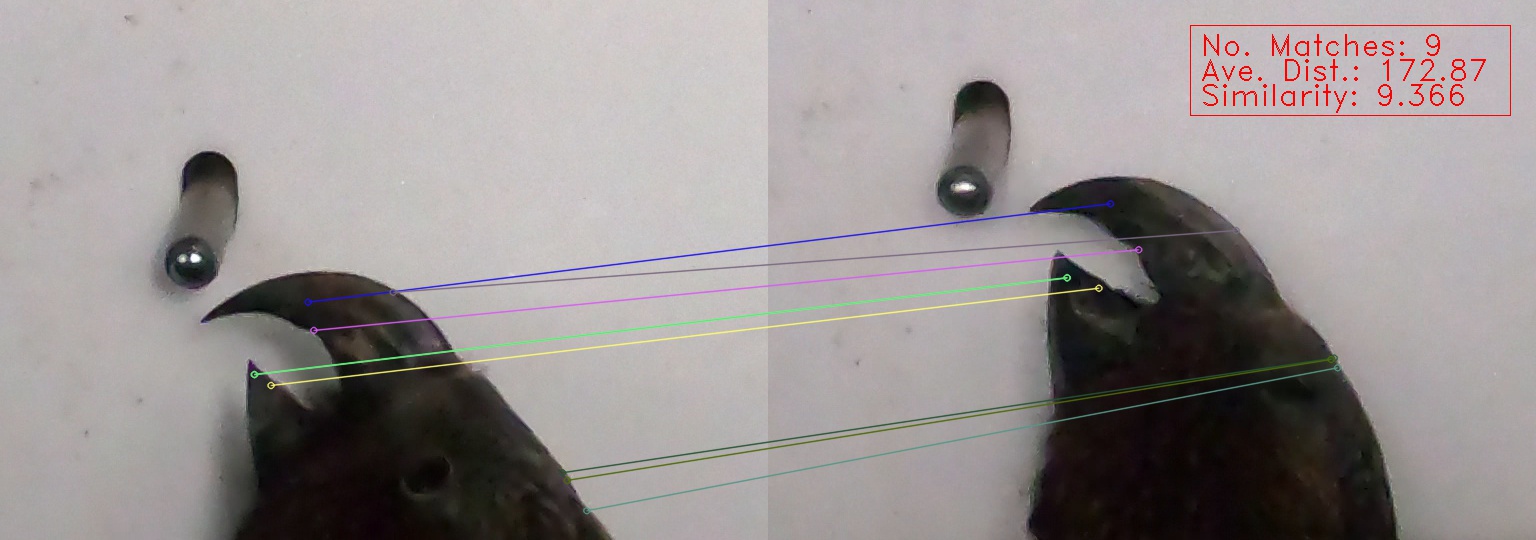}
    \caption{Best image match for \textbf{(g)} in Figure \ref{fig:no-accurate-matches-9}.}
     \label{fig:close-match1}
\end{figure}

\begin{figure}[H]
\centering
  \includegraphics[width=\textwidth]{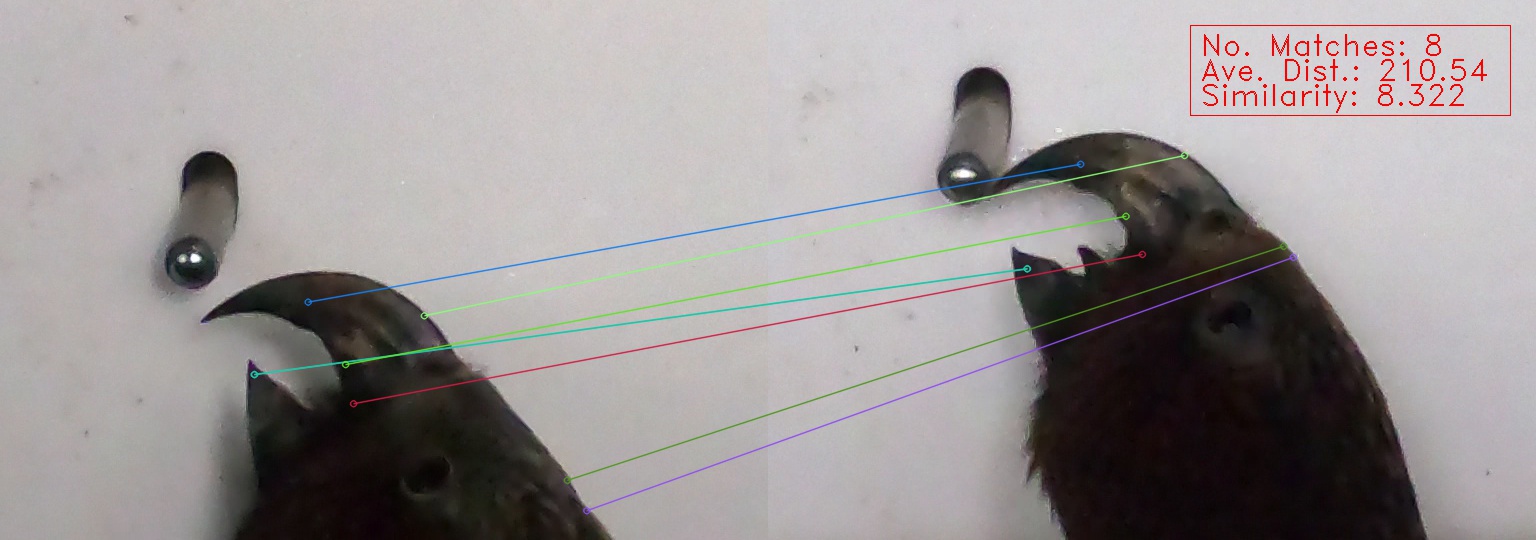}
    \caption{Third best image match for \textbf{(g)} in Figure \ref{fig:no-accurate-matches-9}.}
    \label{fig:close-match2}
\end{figure}

Further analysis of the false matches for the YellowPurple-Yellow band combination show that while the matches were false, visual similarities between images are apparent. As shown in Figures \ref{fig:close-match1} and \ref{fig:close-match2}, the two kākā on the right share similar markings, colouration and geometry of the beak with \textbf{(f)} on the left. Certainly, both matches are false; the kākā matched with in Figure \ref{fig:close-match1} and \ref{fig:close-match2} lack the white ring around the eye present in \textbf{(f)} that indicate that the kākā is a juvenile (young kākā), as well as the lighter marking just above the mouth on the upper mandible of \textbf{(f)}. 

Yet, these are not egregiously incorrect matches and that these are the predicted best matches should perhaps be interpreted positively. We would expect there to be a spectrum of similarity between all images, with the images that contain kākā with no similarity amongst them being the most dissimilar from one another. Kākā that have similar markings on the beak and only differ in minor aspects such as the exact intricacies of the markings and the ring of white around the eye should score higher on the similarity measure, which is what we see here. In terms of incorrect matches, these are more understandably incorrect and reinforce the challenge of the task of differentiating between images of highly similar individuals. 

\begin{figure}[H]
\caption{Sample of images accurately matched with the YellowPurple-Yellow label.} 
\centering 
\begin{tabular}{c c c}
\includegraphics[width=.3\textwidth,valign=m]{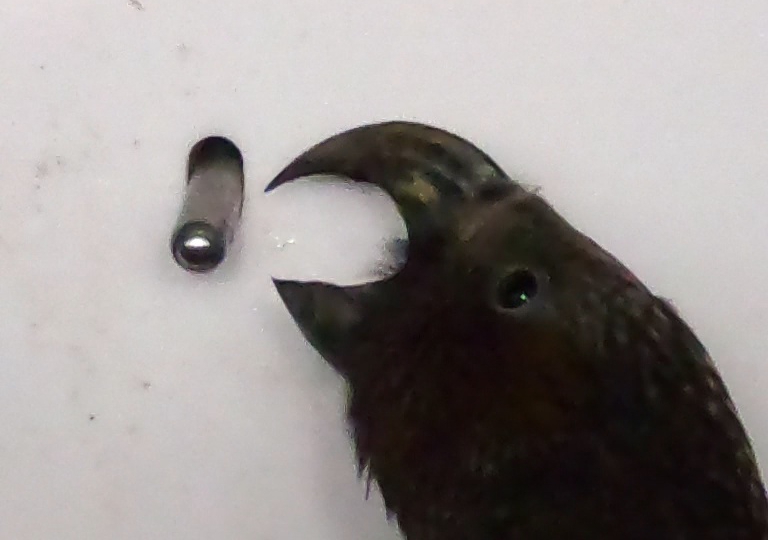} & \includegraphics[width=.3\textwidth,valign=m]{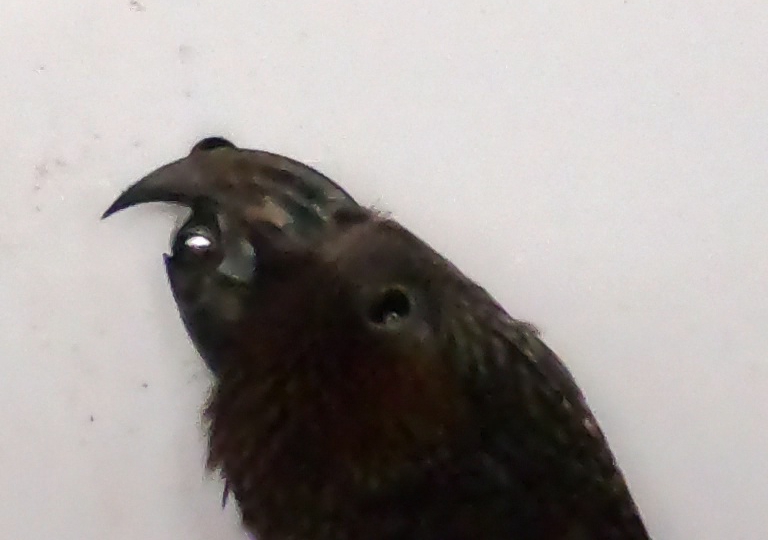}&
\includegraphics[width=.3\textwidth,valign=m]{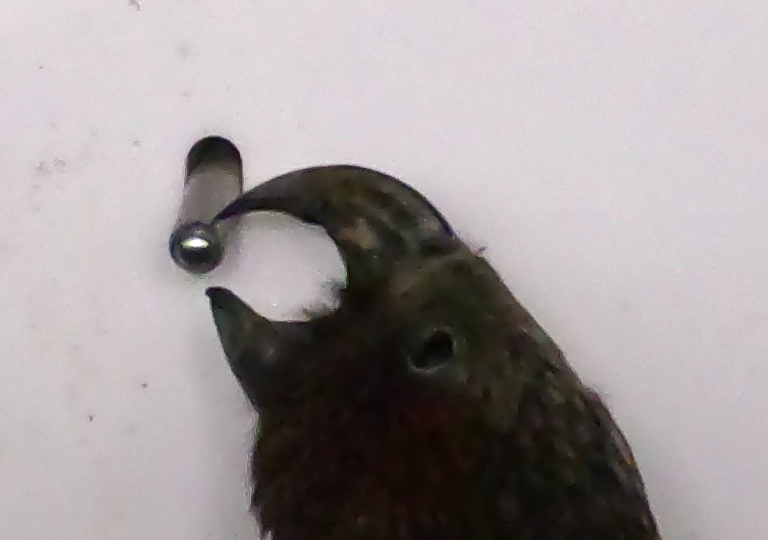} \\
\textbf{(i)} & \textbf{(j)} & \textbf{(k)} \\
\end{tabular}
\label{fig:correct-match-samples}
\end{figure}

When we contrast images of the YellowPurple-Yellow label that were falsely matched with images that were correctly matched, the contrast in quality of the images is evident. A sampling of the correctly classified images is shown in Figure \ref{fig:correct-match-samples}.

In all three of the images pictured in \ref{fig:correct-match-samples} the beak is fully visible in profile to the camera. Even though \textbf{(j)} has the beak slightly obscuring the nozzle of the feeder, the majority of the interesting features on the beak are still visible, and a good view of the region around the eye and top the beak would still be clearly visible after feature detection. \textbf{(i)} represents an ideal input image because the kākā is in the feeding motion, presenting its beak in profile to the camera, and the nozzle would be easily removed during object detection and mask filtering. The image in \textbf{(k)} is also a model input image, even though the nozzle and the kākā overlap, meaning they would not have been separated during image segmentation. The kākā localisation mask for this image might look like Figure \ref{fig:blob-fail}, and illustrates the point that the tips of the beak are not important for feature matching.

\begin{figure}[H]
\centering
  \includegraphics[width=\textwidth]{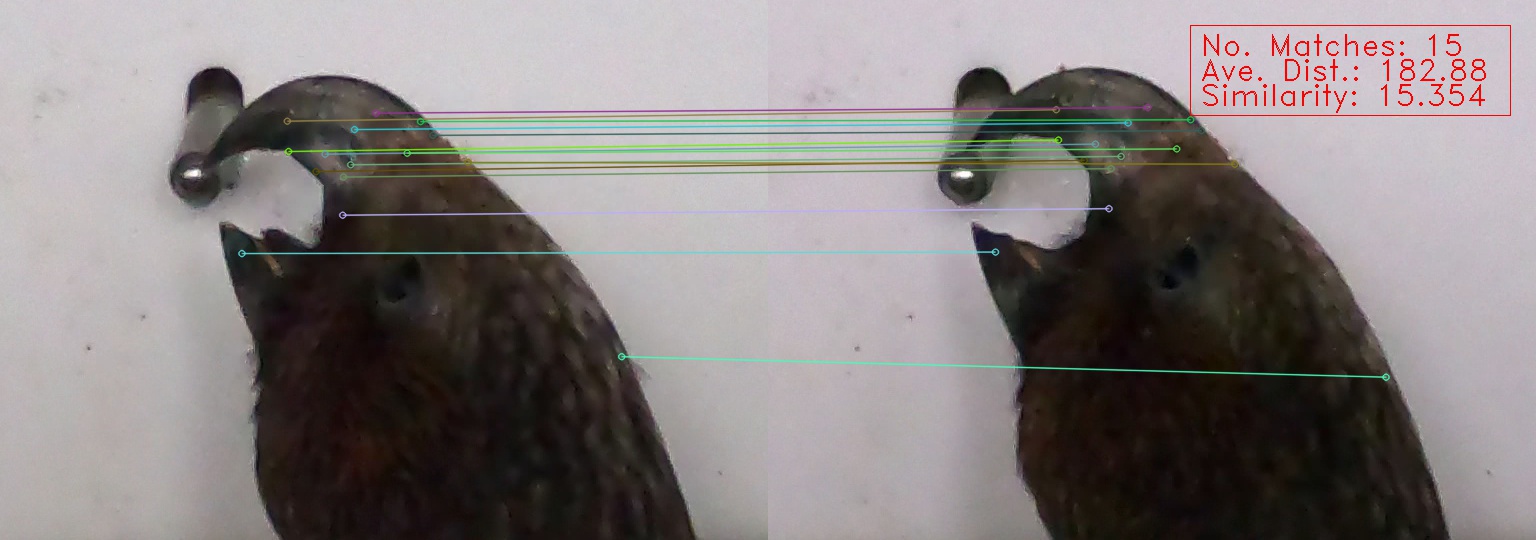}
  \caption{Example of a match classified as false that contains the same individual kākā.}
  \label{fig:falsely-incorrect}
\end{figure}

There are also instances where matches are incorrectly classified as a false match, as hypothesised in Section \ref{sec:labelled-subset}. An example of this is shown in Figure \ref{fig:falsely-incorrect}, where we see the best match identified for an image with the LimePurple-Green label that the image function did not find an accurate match for in the top three best matches. Visually, we can clearly see that the two kākā shown are the same individual bird. The geometry of the lower mandible of the beak, the markings on the beak, and the colouration of the feathers around the nostrils, behind the eye and along the back of the head all match. This would indicate that these two kākā are the same individual.

\section{Summary}

Understanding the margin of error in the accuracy measure and investigating the reasons behind images being falsely matched is important so that we can more accurately evaluate the image matching process. When we account for the nature of the incorrect matches discussed in this section, we understand that a number of the false matches can be reasonably explained by poor quality of input data or obvious visual similarity between kākā. We expect poor quality images to be difficult to match correctly with and likewise we expect there to a high level of visual similarity between kākā. Indeed, it is a reminder of the challenge of differentiating between individuals of the same species and the reason why this research is useful. 

Furthermore, the issues with images that could not find accurate matches are closely connected. The performance of the image matching pipeline is dependent upon the input data, as it requires data of high enough quality that interesting and relevant features can be extracted, and high enough quantity that matches can be found for images that are not ideal image inputs. Certainly, the input image data is one aspect of this process which could be improved upon and affect the performance of image matching positively.

\chapter{Conclusions and Future Work}\label{C:conclusion} 

\section{Conclusions}

This project presented a feature-based image matching pipeline using SIFT feature extraction, MNN feature matching and RANSAC that could ultimately be used to create a similarity network for clustering images and identifying individual kākā. This approach is online and unsupervised where previous work to identify individuals of the same species have used supervised image classification techniques that utilise CNN and other deep learning-based methods as well as vast quantities of labelled data. 

The image matching pipeline was evaluated by calculating the best match for each image from the labelled subset of the data and checking that the best match for each image shares the same label. The number of best matches where each image in the match were of the same label was then used to calculate an overall accuracy score. We also analysed how accuracy was affected by considering the top two best matches and the top three matches to give a more accurate evaluation of how well the image matching pipeline performed.

For images that were inaccurately matched, we observed poor image quality while other false matches were found to be with visually similar images where it would be reasonable to expect that they scored very closely in similarity with the correct matches. Additionally, there were some matches that were false classified as inaccurate due to not all images of the individual of each respective label being labelled. This means that the accuracy of 89.31\% achieved when considering the top three matches is likely an underestimation of the accuracy of the image matching process. From these results, we can conclude that the image matching pipeline achieves a level of accuracy sufficient to effectively construct a similarity network for clustering images and identifying individual kākā. 

Given the challenge of identifying individuals of the same species, this represents considerable progress towards a feature-based similarity network alternative to existing deep-learning based methods.  

\section{Future Work}

Given that the image matching pipeline was developed to be part of a larger system, leaving work to be done, there are many avenues for future work which could be explored using this research as a starting point. 

\begin{itemize}
	\item As concluded in the results from this research, the image matching function is at a stage where it could be used in a preliminary testing capacity with a similarity network to cluster images and identify individual kākā. This could also involve network analysis and experimentation with explosive percolation thresholds as discussed in Section 2.8.
    \item Future work on this pipeline could benefit from improved quality and quantity of data which could be a part of a future summer research project like the one which began this project. Part of the work for this project conducted last summer was to do with iteratively improving the feeder for better data collection (i.e. positioning of the ledges beneath the feeder, adding the matte white feeder background. With this dataset collection method in place, a camera with higher resolution could be used to collect data that would be ready for immediate use in the  This could improve the performance of the existing pipeline and help with any future research endeavours that might build upon this work.
    \item The image extraction process used in this research could be vastly improved upon. For instance, if image extraction were optimised so that images were extracted before it began drinking from the nozzle but when the kākā's beak is in full profile this could avoid situations like those illustrated in {Figure 4.19} where features are lost because superimposing the background mask filters features where the nozzle and beak overlap. Additionally, improving the image extraction method might yield more images from the existing video footage to increase the quantity of data. 
    \item Deep-learning based approaches to feature detection and description are a growing area of research in image matching. This could also be explored in a comparative study, comparing the results from a deep learning based image matching process with this SIFT-based image matching pipeline. 
    \item In a similar vein, other feature extraction methods such as SURF, ORB, BRISK and AKAZE could be explored. Expanding this research in that direction would be a relatively simple to implement but offer an interesting comparison of feature extraction methods in a real-world context. According to the studies cited in justification of using SIFT as the feature extraction method, BRISK performs the second best in accuracy while also being in the top two for efficiency out of SIFT, SURF, ORB, AKAZE and KAZE.
    \item Object detection could be extended to auto-crop images to the beak of kākā rather than simply localising the entire kākā. The majority of the keypoints detected by SIFT were located on the kākā's beak as this is their most distinctive region, so this could improve the results of feature extraction, matching and the overall pipeline. 
    \item The most obvious future direction for this research would be to apply the same image matching pipeline to other species of birds or animals. 
\end{itemize}
\include{formatting}


\backmatter


\bibliographystyle{ieeetr}
\bibliography{bibliography}

\end{document}